\documentclass[letterpaper]{article} 
\usepackage{aaai2026}  
\usepackage{times}  
\usepackage{helvet}  
\usepackage{courier}  
\usepackage[hyphens]{url}  
\usepackage{graphicx} 
\usepackage{natbib}  
\usepackage{caption} 
\usepackage{algorithm}
\usepackage{algorithmic}
\usepackage[labelformat=simple]{subfig}
\usepackage{color,xcolor}
\usepackage{amsmath}
\usepackage{amssymb}
\usepackage{pifont}
\usepackage{array}
\usepackage{multirow}
\usepackage{booktabs}
\usepackage{makecell}
\usepackage{amsthm}

 \usepackage{multirow}
\newtheorem{definition}{Definition}

\usepackage{xcolor}

\newcounter{checksubsection}
\newcounter{checkitem}[checksubsection]

\usepackage{newfloat}
\usepackage{listings}
\DeclareCaptionStyle{ruled}{labelfont=normalfont,labelsep=colon,strut=off} 
\floatstyle{ruled}
\newfloat{listing}{tb}{lst}{}
\floatname{listing}{Listing}
\nocopyright

\title{Think Small, Plan Smart: Minimalist Symbolic Abstraction and Heuristic Subspace Search for LLM-Guided Task Planning}

\author{
\normalfont Junfeng, Tang\textsuperscript{1,2}, 
\normalfont Yuping Yan\textsuperscript{2},
\normalfont Zihan Ye\textsuperscript{3},
\normalfont Zhenshou, Song\textsuperscript{4},
\normalfont Zeqi, Zheng\textsuperscript{1,2}, \\
\normalfont Yaochu Jin\textsuperscript{2}$^\dagger$ \\
\textsuperscript{1}Zhejiang University \quad 
\textsuperscript{2}Westlake University \quad
\textsuperscript{3}UCAS-Terminus AI Lab\quad 
\textsuperscript{4}Victoria University of Wellington\\
\tt\small \{tangjunfeng, Yanyuping, zhengzeqi, jinyaochu\}@westlake.edu.cn, zihhye@outlook.com, zhenshou.song@vuw.ac.nz,
}

\usepackage{bibentry}

\begin{document}

\maketitle

{\renewcommand{\thefootnote}{}
\footnotetext{$^\dagger$ Corresponding author.}}

\begin{abstract}
Reliable task planning is pivotal for achieving long-horizon autonomy in real-world robotic systems. Large language models (LLMs) offer a promising interface for translating complex and ambiguous natural language instructions into actionable plans. However, their probabilistic and opaque nature often leads to logically inconsistent or infeasible outputs. To address these limitations, recent frameworks combine LLMs with symbolic planners by first generating action models (Planning Domain Definition Language) and then applying heuristic search. Although promising, such systems still suffer from representation redundancy and exponential search complexity, often resulting in inefficient or overly long plans. To improve planning efficiency and effectiveness, we propose PLAHX (\textbf{P}lanning from \textbf{L}anguage using \textbf{A}bstraction and \textbf{H}euristic e\textbf{X}ploration), a two-stage LLM-symbolic planning framework that integrates abstract symbolic representations with meta-heuristic subspace search in a parallel and iterative fashion. Rather than relying on verbose LLM-generated domain models, we introduce a minimalist symbolic abstraction pipeline that preserves semantic fidelity while eliminating redundancy. Our approach redefines LLM-symbolic planning not by making LLMs smarter, but by reducing the symbolic search space adaptively. Empirical results across four challenging domains, including block stacking and robotic mobile grasping, show that our approach improves the success rate by 21.47\% on average, while reducing token consumption by 13\% compared to state-of-the-art baselines.

\end{abstract}


\section{Introduction}

\begin{figure}[h]
    \centering
    \includegraphics[width=1\linewidth]{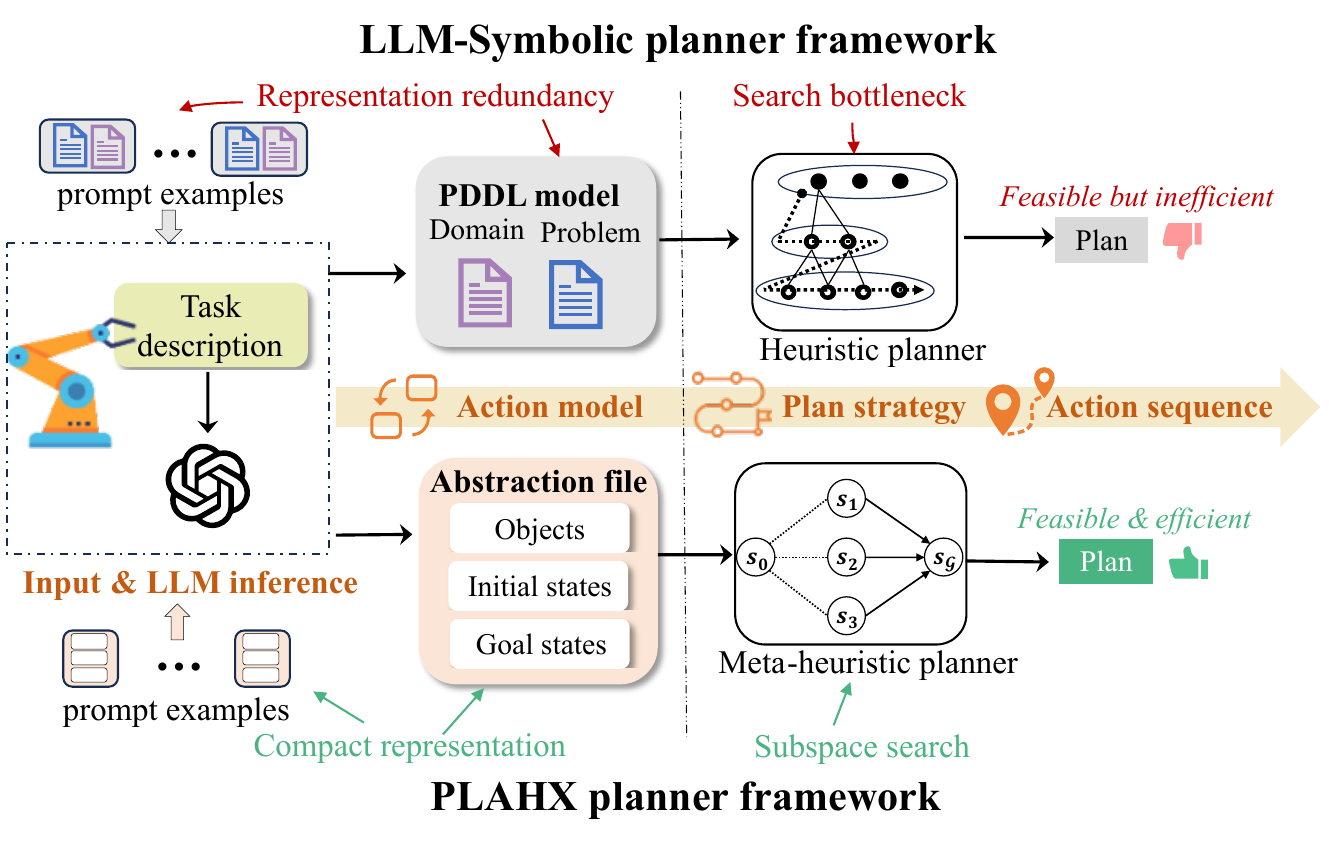}
    \caption{Comparison between prior LLM-symbolic planning frameworks and the proposed PLAHX framework.}
    \label{fig:banner}
\end{figure}

Large language models (LLMs) have emerged as powerful “robotic brains” for task planning, offering strong capabilities in language understanding, reasoning, and generalization \cite{kannan2024smart, Song_2023_ICCV, de2024large}. They have shown promise in embodied domains such as navigation \cite{latif20243p} and household manipulation \cite{glocker2025llm}. Despite their versatility, LLMs can function as probabilistic black-boxes due to the lack of explicit structure and reasoning guarantees, often leading to unreliable behaviors in long-horizon tasks that require precise symbolic reasoning, state tracking, and temporal coherence. This uncertainty in decision-making can lead to critical failures in downstream execution. This is particularly significant in domains where robust and interpretable plans are essential, such as route planning and multi-step robotic actions.

To improve the reliability of LLM-guided planning, recent work has focused on integrating LLMs with symbolic planners, where LLMs interpret task descriptions and generate symbolic representations, while the symbolic planner ensures verifiable and goal-directed search \cite{DBLP:journals/corr/abs-2503-18971, silver2022pddl, liu2023llm+p, chu2025llm+map}. Typically, the planning problem is symbolized using the Planning Domain Definition Language (PDDL) \cite{IntroPDDL2020} that involves domain files (defining predicates and action schemas) and problem files (specifying initial states and goals). A domain-independent planner then applies heuristic search algorithms \cite{Segovia-Aguas_Jiménez_Jonsson_2021} to compute a feasible plan. These methods have been proven effective in improving the success rate and explainability of robotic task planning.

Despite recent progress, current LLM-symbolic planning approaches face two fundamental limitations that hinder their ability to generate plans efficiently, as shown in Figure \ref{fig:banner}. (1) \textbf{Representation redundancy}: to enable LLMs to generate complete PDDL models, prompts are often overloaded with full domain–problem pairs. This quickly exhausts the LLM’s context window and incurs significant token overhead. (2) \textbf{Exponential search bottleneck}: once the PDDL model is constructed, most approaches rely on heuristic planning methods (e.g., A$^*$, GBFS, Fast-Downward) \cite{AStar2025Robotic,LLMheuristic2025,FastDownward2006} to generate executable paths. Although the heuristic computation is polynomial in time, the grounded state–action space grows exponentially with the number of objects, predicates, and action schemas, making efficient search increasingly intractable and often resulting in suboptimal plan lengths \cite{ReviewOfAutomatedPlanning}.

To mitigate the aforementioned limitations and enable effective and efficient planing, we propose a method called \textbf{P}lanning from \textbf{L}anguage using \textbf{A}bstraction and \textbf{H}euristic e\textbf{X}ploration (PLAHX). As shown in Figure \ref{fig:banner}, instead of full symbolic representation and heuristic planning, the proposed two-stage LLM-symbolic planning framework integrates compact symbolic abstraction with meta-heuristic subspace search. 
Our contributions are threefold:
\begin{itemize}
    \item We introduce a minimalist symbolic abstraction approach in the translation stage, which reduces LLM token consumption while preserving domain semantics, by extracting only the initial and goal states instead of full domain–problem pairs, thereby mitigating context overflow and domain drift.
    \item We propose a hybrid meta-heuristic planner in the planning stage that combines global diversity and local efficiency: a population-based mechanism maintains diverse action subspaces, and heuristic search within each subspace yields feasible and length-optimal plans.
    \item We empirically demonstrate that PLAHX significantly outperforms existing LLM-symbolic baselines in both planning success rate and efficiency across four challenging robotic domains, achieving 21.47\% success rate improvements.
\end{itemize}

\section{Related Work}
The emergence of LLMs as a task-agnostic reasoning module presents a promising pathway to general robot planning capabilities. These planning approaches can be divided into two categories in terms of whether supporting formal representation: directly LLM-guided planning and hybrid LLM-symbolic planning.
\subsection{Directly LLM-Guided Planning}
LLMs have been increasingly applied to map natural language instructions to actionable plans in robotic task planning \cite{huang2022language,Song_2023_ICCV,brohan2023can,de2024large}. The study \cite{huang2022language} was among the first to use prompt engineering to directly generate executable action plans from natural language inputs. Subsequent research has focused on improving the generalizability, success rate, and computational efficiency of LLM-guided planning systems. For example, LLM-Planner \cite{Song_2023_ICCV} introduces a few-shot planning framework that reduces data collection costs and improves sample efficiency, enabling robots to follow complex language instructions. SayCan \cite{brohan2023can} uses an action selection mechanism that evaluates affordances through learned value functions, improving the decision-making process in robotic systems. Furthermore, recent work \cite{de2024large} has employed prompting techniques to guide LLMs in converting high-level language instructions into admissible actions in the domain of robotic dance creation, demonstrating the potential of LLMs in accomplishing creative and complex tasks.

\subsection{Hybrid LLM-Symbolic Planning}
Despite these advances, recent studies have shown that LLM-guided planners often produce incorrect or infeasible actions, particularly in long-horizon tasks \cite{valmeekam2022large,valmeekam2023planning}. As emphasized in \cite{PlanDark2025HuangLC25,kambhampati2024position}, a significant challenge in planning and reasoning are predominantly linked to thinking, characterized by slow, deliberate, and conscious cognitive processes \cite{sloman1996empirical,kahneman2011thinking}. The behaviors of LLMs being in line with no first-principle reasoning, that exhibit constant response times regardless of the complexity of the questions posed. there is a growing interest in exploring hybrid approaches. For instance, LLM+P \cite{liu2023llm+p} relies on language understanding capabilities of LLMs to generate symbolic models from task descriptions, while leveraging the formal structure and interpretability of PDDL to ensure plan validity. The approaches we focus on in this work first utilize LLMs in translating natural language to symbolic models, and then resort to an external symbolic approaches to produce plans or assist validation.
LLM+MAP \cite{chu2025llm+map} generates a PDDL representations from language descriptions, and then produce plans.
ISR-LLM \cite{zhou2024isr} introduces an iterative self-refined procedure where the LLM repeatedly updates the plan in coordination with the PDDL model. While exist studies have proposed a mature pipeline to solve domain-specific problems, using LLMs to construct PDDL representations from languages and employing heuristic planners, they overlooks two critical challenges: domain drift \cite{DomainDrift2024}, which introduces inconsistencies or omissions in symbolic representations, and the computational cost in searching. In contrast, our work strives to reduce representation redundancy, circumventing the unnecessary symbolics, and adopt parallel accelerate search to provide a more resource-saving approach to hybrid LLM-symbolic planning.

\section{Preliminaries}
We review the Planning Domain Definition Language (PDDL) first, a foundational framework for specifying planning problems. We then formulate the limitations inherent in current LLM-guided task planning with PDDL, highlighting key challenges that necessitate investigation in this work. Finally, we introduce the formulation of task planning.
\label{sec:preliminary}
\subsection{Planning Domain Definition Language}
PDDL is a standardized formalism widely used for representing planning problems in automated planning systems \cite{DBLP:journals/aim/McDermott00,PDDL2003,IntroPDDL2020}. A PDDL specification typically consists of two files: (1) \textit{a domain file}, which defines the object types, predicates and action schemas, and (2) \textit{a problem file}, which specifies the planning task by listing the objects involved, the initial state, and the goal conditions. Each action schema encodes an operator’s parameters, preconditions, and effects, and can be instantiated (i.e., grounded) by substituting concrete objects for parameters. Its preconditions and effects are a set of states. The initial states and goals of planning problems are a set of atoms with predicates instantiated from objects. Example domain and problem files used in our experiments are provided in Appendix A.

\subsection{Problem Formulation}
We formally characterize two fundamental limitations of LLM-based task planning with PDDL: (1) representation verbosity, which leads to context overflow and domain drift; and (2) exponential bottlenecks, which result in prohibitive time complexity during symbolic search.
\subsubsection{Limitation I: Representation Verbosity.}

Let  
\begin{itemize}
  \item $\mathcal{P}$ denote the set of prompts, each $p \in \mathcal{P}$ consisting of $k$ full PDDL domain–problem pairs concatenated together,
  \item $\ell_{\max}$ be the LLM’s maximum context length in tokens,
  \item $\mathcal{C}$ indicate the conversation completion via LLMs,
  \item $D_{\Theta}$ be the PDDL domain file generated by the LLM with $\Theta$ representing its set of solvable planning problems.
\end{itemize}

\begin{definition}
    Context overflow occurs when the total number of tokens $t$ consumed by the prompt and its completion exceeds the model’s maximum context length $\ell_{\max}$:

  \begin{equation}
    \textsc{Overflow}\triangleq t(\mathcal{P})+t(\mathcal{C}(\mathcal{P}))>\ell_{\max}.
    \label{eq:overflow}
  \end{equation}
\end{definition}

\begin{definition}
    Referenced to ``heuristic domain equivalence'' \cite{DBLP:conf/icaps/OswaldSK00S24},
    given a original planning domain $D_{\Theta}$, and a generated planning domain ${D_{\Theta}}^{\prime}$, domain drift is defined as:
    \begin{equation}
    \textsc{Drift}(\Theta,\Theta^{\prime})=\frac{|\Theta\setminus\Theta^{\prime}|}{|\Theta|}.
    \label{eq:drift}
  \end{equation}
  $|\Theta\setminus\Theta^{\prime}|$ means the number that each problem $\theta \in \Theta$ cannot be transformed into the problem set $\Theta^\prime$ undering domain ${D_{\Theta}}^{\prime}$.
  If $\textsc{Drift}(\Theta, \Theta^{\prime}) > \tau_{d}$, where $\tau_{d}$ is a predefined threshold, we consider that domain drift has occurred.
\end{definition}

\subsubsection{Limitation II: Exponential Bottleneck.} Let 
\begin{itemize}
    \item $S$ be the set of atoms with predicates from $P$ instantiated with objects from $O$.
\end{itemize}
\begin{definition}
    The complexity of deterministic search over the symbolic space satisfies:
\begin{equation}
  T_{\text{search}} = \Omega\bigl(|S|^{1 - \gamma}\bigr), \quad \gamma \approx 0,
  \label{eq:bottleneck}
\end{equation}
where $|S| \propto |P|\times|O|^q$ scales exponentially where $q$ is the number of parameters of the predicate.
This exponential scaling renders exact planning increasingly intractable for large or complex domains.
\end{definition}

\begin{figure*}[ht]
    \centering
    \includegraphics[width=0.76\linewidth]{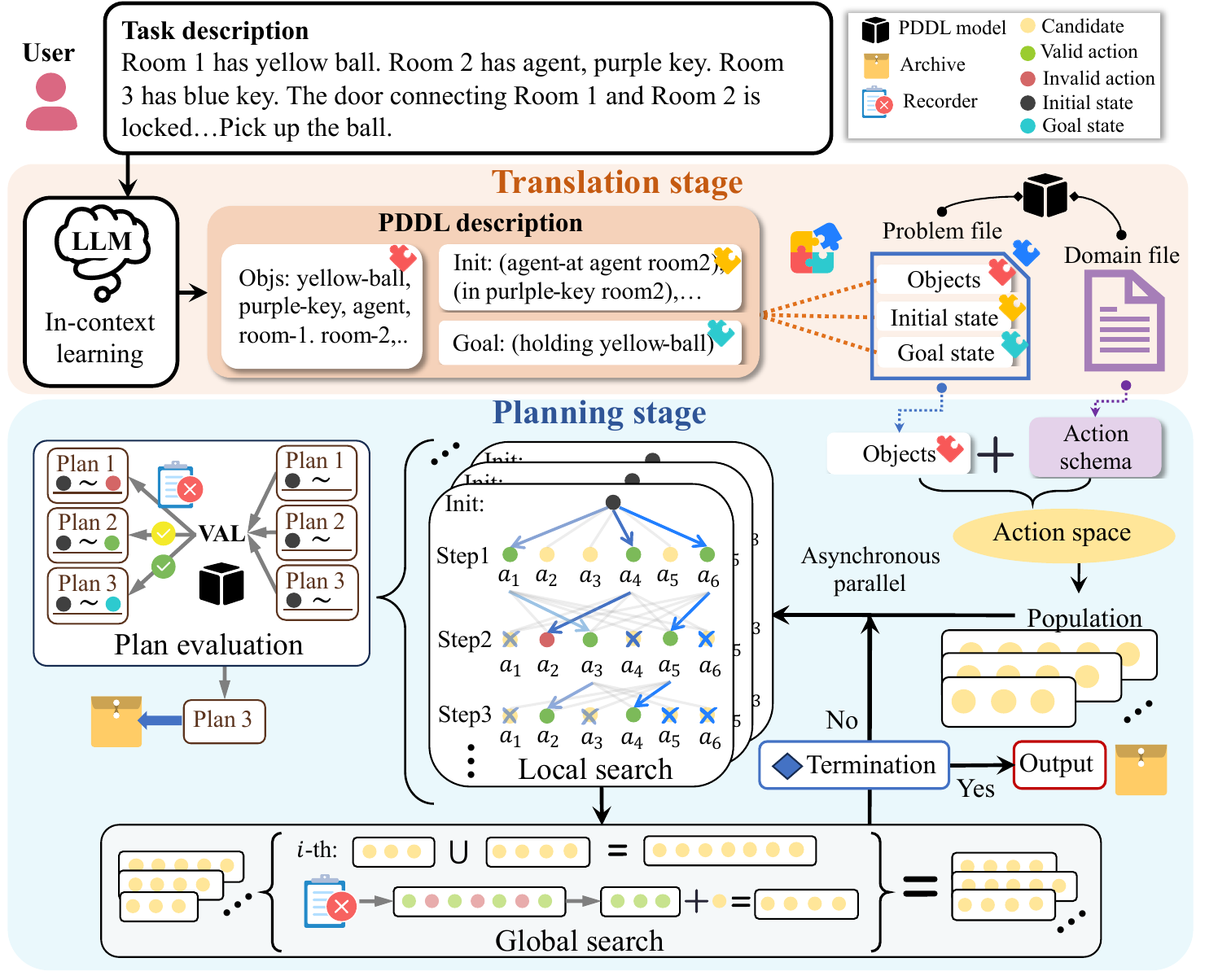}
    \caption{Overview of the proposed PLAHX, which decomposes the language-guided planning process into two stages. Given a task description, PLAHX first translates natural languages into symbolic representations for formal validation. Subsequently, a population-based search mechanism intertwining global and local search is employed to obtain the plan effectively and efficiently.}
    \label{fig:overview}
\end{figure*}

\subsection{Task Formulation}
We study deterministic, fully observable task planning problems formulated in the classical STRIPS (STanford Research Institute Problem Solver) subset of PDDL \cite{DBLP:conf/ijcai/FikesN71}. Each planning task is defined as a tuple:
\begin{equation}
    Q = \langle S, A, \mathcal{T}_f, s_\mathcal{I}, s_{\mathcal{G}} \rangle,
\end{equation}
where $S$ is a discrete set of states, $A$ is a set of action schemas, $\mathcal{T}_f: S \times A \rightarrow S$ is the transition function that defines state transitions, $s_\mathcal{I} \in S$ is the initial state, and $s_{\mathcal{G}} \in S$ denotes the goal condition(s).

For any state $s \in S$, an applicable action $a \in A(s) \subseteq A$ can be selected if its preconditions are satisfied. An action sequence (or plan) $\pi = (a_1, \ldots,a_n)$ is a valid solution if it transforms the initial state $s_\mathcal{I}$ to a goal state $s_{\mathcal{G}}$ via successive applications of $\mathcal{T}_f$. Each $a_i$ is a grounded instance of an action schema $\alpha \in A$, instantiated by substituting parameters with concrete objects from a finite set $O$.

For long-horizon sequential planning tasks, the plan length $n$ is typically large, and the state space $|S|$ expands exponentially with the number of objects, predicates, and action schemas, approximately as $|S| \propto {|P|\times |O|^q}$. In PDDL-based models, the domain file specifies the transition dynamics $\mathcal{T}_f$ through action schemas, while the problem file defines the object set $O$, initial state $s_\mathcal{I}$, and goal specification $s_\mathcal{G}$. In this work, we focus on the LLM-symbolic method as a planner to solve domain-specific planning problems.

\section{PLAHX Framework}
\label{sec:approach}
To generate actionable plans effectively and efficiently from natural language instructions, PLAHX adopts a two-stage pipeline that jointly addresses representation redundancy and the exponential search bottleneck. As shown in Figure~\ref{fig:overview}, the framework consists of a translation stage and a planning stage. Throughout this process, we assume the domain file and the structural format of the problem file are predefined, allowing PLAHX to focus on efficient problem instantiation and plan generation within a known symbolic domain.

\subsection{Phase 1: Translation Stage}
\label{sec:translation}
In the translation stage, PLAHX employs an LLM-based translator to synthesize a minimal symbolic representation that captures the essential planning semantics while staying within the LLM's context budget. Given an open-ended task description, the translator extracts the object set, initial state, and goal state $Z = \langle O, s_\mathcal{I}, s_{\mathcal{G}} \rangle$ where the initial state $s_\mathcal{I}$ and goal $s_{\mathcal{G}}$ are set of atoms. 

\subsubsection{Prompt Construction}
To encode the language in the desired symbolic representation, we utilize in-context learning by prompting the LLM translator with few-shot demonstrations: 
  \begin{equation}
    \mathcal{P}= [u_1, z_1, \ldots, u_k, z_k],
  \end{equation}
where $u_i$ denotes a natural language task description and $z_i$ is the corresponding abstractions. The final query $u_{query}$ represents the target task to be translated. By conditioning the LLM translator on this structured prompt, we effectively cast symbolic translation as a conditional sequence prediction task, where the LLM completes $u_{query}$ with a predicted symbolic output $z_{pred} \in \mathcal{P}$.

\subsubsection{Formal Guarantee}
As $Z = \langle O, s_\mathcal{I}, s_{\mathcal{G}} \rangle$ contains the objects, the initial state and goal atoms, the formal soundness of this representation depends on accurate grounding and complete coverage of the required predicates and object references. To enable plan validation in the subsequent stage, $Z = \langle O, s_\mathcal{I}, s_{\mathcal{G}} \rangle$ is embedded in a PDDL problem file, which is then paired with a predefined domain file to form a complete PDDL model. This model is passed to VAL \cite{VAL2004}, a standard plan validator, to ensure logical consistency and execution feasibility.

\subsection{Phase 2: Planning Stage}
\label{sec:planning}
To reduce search of unnecessary actions, PLAHX employs a meta-heuristic planner that maintains a population-based search in the ground action space. The planner partitions the grounded action space $\mathcal{A}$ into $K$ lightweight subspaces, where each individual represents a candidate subset of actions for localized heuristic search. This structure enables scalable and parallel exploration of large state spaces.

\subsubsection{Action Grounding}
\label{sec:grounding}

Given the symbolic representation of a task description $Z = \langle O, s_\mathcal{I}, s_{\mathcal{G}} \rangle$, the planner instantiates all action schemas $\alpha \in A$ with valid combinations of objects from $O$. The resulting grounded action set $\mathcal{A} = \{a_1, \ldots, a_N\}$ is linearized and indexed to support subspace construction and manipulation during the search. 

\subsubsection{Meta-Heuristic Population Search}
\label{sec:meta-search}
Due to the exponential growth of $\mathcal{A}$, exhaustive search is prohibitively costly and impractical. We adopt a meta-heuristic approach inspired by genetic algorithms,  where the action space is partitioned into diverse subspaces via a population-based mechanism and conducts parallel local search within each one to reduce runtime.

\paragraph{Encoding Process.}
An individual $\mathcal{A}^i$ is encoded as a variable-length integer sequence: 
  \begin{equation}
    \mathbf{x}^{(i)} = (a^{(i)}_{1}, \ldots, a^{(i)}_{l_i}), \quad 3 \le l_i \le  \left\lfloor |\mathcal{A}|/2\right\rfloor,
    \label{eq:encoding}
  \end{equation}
where $l_i$ is the individual length. It implicitly defines a lightweight action subspace $\mathcal{A}^{(i)} \subset \mathcal{A}$.
The population contains multiple individuals, which partitions the overall (ground) action space into multiple subspaces, explicitly reducing the number of search nodes. Next, local search is employed on each individual.

\paragraph{Local Search Process.}
Each individual $\mathbf{x}^{(i)}$ undergoes an asynchronous \textit{A}$^*$ search \cite{hart1968formal} in its subspace $\mathcal{A}^{(i)}$. The algorithm maintains an open list of partial plans $\pi = (a_1,\ldots,a_k)$, prioritized by the estimated cost:
  \begin{equation}
    f(\pi) = g(\pi) + h(\pi),
    \label{eq:estimated_cost}
  \end{equation}
where $g(\pi)$ denotes the exact accumulated cost and $h(\pi)$ is ``fast-forward heuristic'' \cite{FastForward2001,FastForward2011} used to quickly estimate the distance from the current state to the goal.

During the search process, we maintain a global cost threshold $f_{\text{global}}$. A node is expanded only if its estimated cost $f(\pi)$ is less than $f_{\text{global}}$, which effectively prunes unpromising branches and guides the search toward the more viable paths. Whenever a local search discovers a plan $\pi$ with a lower estimated cost, the global estimated cost is updated to $f_{\text{global}} = f(\pi)$.

\paragraph{Plan Evaluation.}
During each local search, the VAL validator \cite{VAL2004} is invoked to assess candidate plans based on two criteria: 
\begin{enumerate}
  \item \emph{Goal test:} check whether the current state satisfies the goal condition $s_{\mathcal{G}}$;
  \item  \emph{Precondition test:} detect any action $a_i$ in the plan $\pi$ has preconditions violates by the effects of a preceding action $a_k$ with $k \leq i$.

\end{enumerate}
These evaluations yield one of three outcomes: (i) the plan succeeds; (ii) the goal condition is unsatisfied; and (iii) an unsatisfied precondition is encountered, indicating a conflict between two actions. To prevent conflicting actions from co-occurring, we maintain a \textbf{conflict recorder} that tracks the pairwise action compatibility. When outcome (iii) is encountered, the compatibility score between the conflicting actions $(a_i, a_k)$ is penalized as follows:
\begin{equation}
\begin{split}
w &(a_i,a_k)  \leftarrow 
                        \max \Bigl(w(a_i,a_k)-\rho\frac{\tau}{\tau-\varepsilon(a_i,a_k)}, \ell_b\Bigr),
 \label{eq:penalty}
 \end{split}
\end{equation}
where $\rho>0$ and $\tau>0$ control the penalty magnitude and decay rate,  $\varepsilon(a_i,a_k)$ is the conflict count incremented upon each failure, and $\ell_b$ is a lower bound ensuring non-negative compatibility scores. By contrast, if the plan succeeds (i), it is archived and preserved for the remainder of the search process.

\paragraph{Global Search Process.}
At each generation, after completing all local searches, we apply the following three-step global search procedure to each individual $\mathbf{x}^{(i)}$ to refine the search space and promote diversity:

\begin{enumerate}
    \item \emph{Pairwise union.} A partner individual $\mathbf{x}^{(j)}$ 
 is randomly selected, and their corresponding actions subspaces are merged to form a union set:
 \begin{equation}
     C = \mathcal{A}^{(i)} \cup \mathcal{A}^{(j)}.
 \end{equation}
    \item \emph{Compatibility-guided sampling.} 
    For each action $a \in C$, its compatibility weight within the union $C$ is computed as: 
\begin{equation}
\omega(a)=\sum_{a^{\prime} \in C \backslash\{a\}} w\left(a, a^{\prime}\right),
\end{equation}
    which measures its executability in the current content $C$. These weights are then normalized to define a sampling distribution:
\begin{equation}
p(a)=\frac{\omega(a)}{\sum_{a^{\prime} \in C} \omega\left(a^{\prime}\right)} .
\end{equation}
Using this distribution, we sample $k$ actions without replacement to construct an intermediate individual $\mathbf{x}^{\prime}$, where:
\begin{equation}
    \min(|\mathcal{A}^{(i)}|, |\mathcal{A}^{(j)}|) \le k \le \max(|\mathcal{A}^{(i)}|, |\mathcal{A}^{(j)}|).
\end{equation}

    \item \emph{Mutation addition.} With a mutation probability $p_{\text{mut}}$, an additional action $a_{new} \in \mathcal{A} \setminus C $ is sampled at a probability proportional to $\omega(a_\text{new})$ and appended to $\mathbf{x}^\prime$. The resulting offspring is:
    \begin{equation}
        \mathbf{x}_\text{new} = \mathbf{x}^\prime\oplus(a_\text{new}), \text{where}\; \mathbf{x}^\prime| \le |\mathbf{x}_\text{new}| \le |\mathbf{x}^\prime|+1.
    \end{equation}

\end{enumerate}

Throughout the global search, any action deemed \emph{critical}, i.e.\, one whose effects directly contribute to the goal or that is executable in the initial state, is preserved in all subspaces to maintain the completeness of the search.

\paragraph{Termination Condition.}
Every plan validated by VAL as successful is stored in an external archive $\prod^*$. The search stops once $|\prod^*|$ reaches a user-defined threshold. Otherwise, the algorithm continues until a maximum number of iterations is reached, after which all archived plans in $\prod^*$ are returned. The final output is the plan $\pi^*$ with the lowest estimated cost $f(\pi^*)$, representing the optimal action sequence.

\section{Experiment}
\label{sec:experiments}
This section evaluates the effectiveness of PLAHX through three core research questions:
\begin{itemize}
    \item \textbf{RQ1: Planning Effectiveness.} Does PLAHX achieve state-of-the-art performance in language-guided planning tasks, particularly in addressing the challenges posed by open-ended language instructions that increase the difficulty of generating actionable plans and the complexity?
    \item \textbf{RQ2: Representation Robustness.} Can the symbolic abstraction strategy mitigate context overflow and domain drift issues in LLM-based symbolic planning?
    \item \textbf{RQ3: Search Efficiency.} Does the meta-heuristic planner effectively overcome the exponential search bottleneck compared to classical heuristic search methods in LLM-based robotic planning tasks?
\end{itemize}

We benchmark PLAHX across four widely studied robotic planning domains: Blocks World, Tower of Hanoi, Grippers, and Rearrangement, all of which are standard testbeds in automated planning research \cite{silver2020pddlgym}. For each domain, we automatically generate over 100 diverse task instances. Each instance includes (i) a natural language instruction, (ii) a corresponding PDDL problem specification, and (iii) the requisite domain model. Additional examples and domain details are provided in Appendix A.

\subsection{Baselines}
\label{subsec:baselines}
At the level of symbolic representation, we evaluate PLAHX against other LLM-symbolic planners with the full PDDL model, problem file and symbolic abstraction. At the level of planning over actions, we evaluate PLAHX with classical heuristic planning methods and an LLM-guide planner. Therefore, we evaluate four distinct methods: (1) LLM-P-FD,  which is the baseline approach grounded in \cite{liu2023llm+p,chu2025llm+map}. It generates problem file and delegates planning to Fast-Downward ~\cite{FastDownward2006}. (2) LLM-P-FF, which invokes the classical Fast-Forward ~\cite{FastForward2001,FastForward2011} to obtain plans. (3) ISR-LLM~\cite{zhou2024isr}, which introduces iterative self-refined LLMs to generate the PDDL model(domain and problem files) and plans. (4) Our proposed PLAHX, which generates symbolic abstractions from language descriptions and employs meta-heuristic planning methods to search plans.

\subsection{Experimental Setup}
\label{sec:set-up}
\subsubsection{Evaluation Metrics}
To evaluate PLAHX, we use a set of metrics corresponding to the three research questions:
\begin{itemize}
    \item \textbf{Success rate}: The percentage of planning tasks, of which a valid and executable plan is successfully generated and verified by the VAL validator.
    \item \textbf{Context overflow:} indicates whether the combined token count of the prompt and the LLM completion exceeds the context length limits, defined as $t(\mathcal{P})+t(\mathcal{C}(\mathcal{P}))>\ell_{\max}(=5000)$.
    \item \textbf{Domain drift:} measures the deviation between the LLM-generated domain and the original, defined as $\textsc{Drift}(\Theta,\Theta_{\text{ex}})>\tau_{d}(=0.01)$.
    \item \textbf{Average token cost $\bar{T}_D$:} quantifies the degree of redundancy, defined as 
  \begin{equation}
    \overline{T}_{\mathcal{D}} \triangleq \frac{1}{m}\sum_{i=1}^{m} T_i,
  \label{eq:avg-token}
  \end{equation}
  where given the domain $D$ and its set of planning problems $\Theta=\{\theta_1,\ldots,\theta_m\}$, $T_i$ is the total number of tokens consumed in the prompt when solving $\theta_i$. A lower value implies more efficient use of context length.
  \item \textbf{Effective search space $|\bar{\mathcal{A}}^{(i)}|$}: the number of grounded actions actually explored during the search, reflecting how well the planner reduces the combinatorial complexity.
  \item \textbf{CPU time (seconds)}: the total runtime required to find and validate a correct plan for each task instance.
\end{itemize}

More details of the experimental setup can be found in Appendix B.

\subsection{Results}
\label{subsec:results}

\subsubsection{Performance Analysis}

\begin{table}[]
\setlength{\tabcolsep}{0.45mm}
\centering
\begin{tabular}{ccccc}
\toprule
Domain               & LLM-P-FF    & LLM-P-FD     & ISR-LLM       & \makecell{PLAHX\\(\textbf{Ours})}               \\ \midrule
Blocks World        & 20.53     & 20.53     & 0             & \textbf{24.50}            \\
Tower of Hanoi      & {82.29}     & \textbf{82.37}     & 0             & 81.14          \\
Grippers             & 72.64     & 72.64     & 0             & \textbf{75.65}          \\ 
Rearrangement      & 84.87     & 84.81     & 50.86         & \textbf{87.85}          \\ 
\midrule
Total               & 76.83     & 77.10   &   16.97         & \textbf{78.44}         \\ 
\bottomrule
\end{tabular}
\caption{The planning success rate (\%) on the test problem across different domains.}
\label{tab:sr}
\end{table}

\begin{figure}[ht]
    \centering
    \includegraphics[width=0.95\linewidth]{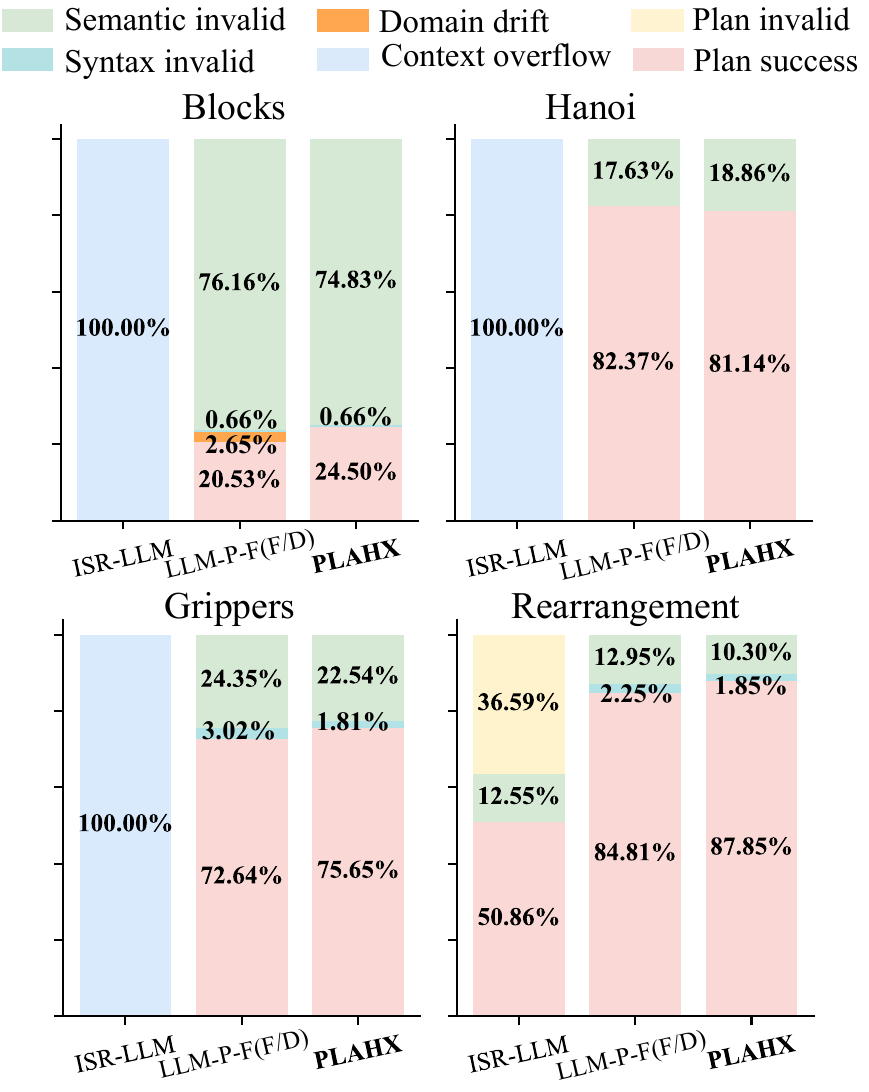}
    \caption{Classes of the results by different LLM-symbolic planners in the considered domains. They are (1) syntax error: LLM output cannot conform to standard PDDL syntax; (2) semantic error: the output is syntactically valid but inconsistent with task descriptions; (3) plan invalidity: the planner fails to find valid plans to achieve the goal; (4) plan success: the generated plan succeed; (5) context overflow, and (6) domain drift.}
    \label{fig:rc}
\end{figure}

\begin{table}[]
\setlength{\tabcolsep}{0.5mm}
\centering
\begin{tabular}{cccc}
\toprule
Domain               & LLM-P-F(F/D)         & ISR-LLM       & \makecell{PLAHX(\textbf{Ours})}               \\ \midrule
Blocks World         & $1.21$          & $8.16$       & $\textbf{1.08}$($\downarrow$11\%)            \\
Tower of Hanoi       & $1.34$          & $9.92$       & $\textbf{1.21}$($\downarrow$10\%)         \\
Grippers             & $1.79$          & $9.93$       & $\textbf{1.55}$($\downarrow$13\%)         \\ 
Rearrangement           & $1.88$         & $2.78$       & \textbf{1.78}($\downarrow$5\%)          \\ 
\bottomrule
\end{tabular}
\caption{The average token cost ($\times 10^{3}$) on test problems across different domains. It can be observed that PLAHX consumes the fewest tokens, demonstrating superior efficiency and reduce 5-13\% token compared to the second efficient planning method. Notably, LLM-P-FF and LLM-P-FD, which employ the identical translation method, exhibit the same consumption and are collectively denoted as LLM-P-F(F/D).}
\label{tab:atc}
\end{table}

\paragraph{RQ1: Planning Effectiveness.}
The experimental results, as detailed in Table \ref{tab:sr}, highlight the superior performance of PLAHX across tested domains. PLAHX surpasses the baseline methods in three of four domains, with notable gains in Blocks World and Rearrangement. In Blocks World, 24.50\% success rate of PLAHX significantly exceeds LLM-P-FF (20.53\%) and LLM-P-FD (20.53\%), indicating robustness in logically complex environments. In Rearrangement, PLAHX achieves a success rate of 87.85\%, outperforming LLM-P-FF (84.87\%) and LLM-P-FD (84.81\%), demonstrating its effectiveness in detailed planning scenarios.

PLAHX also shows a competitive performance in the Grippers domain with a success rate of 75.65\% , slightly better than LLM-P-FF (72.64\%) and LLM-P-FD (72.64\%). In Tower of Hanoi, PLAHX achieves a success rate of 81.14\%, comparable to LLM-P-FF (82.29\%) and LLM-P-FD (82.37\%), showcasing its capability in complex, recursive tasks.

To summarize, PLAHX attains the highest overall success rate of 78.44\% across all domains, outperforming LLM-P-FF (76.83\%), LLM-P-FD (77.10\%), and ISR-LLM (16.97\%) at a large margin. These results underscore PLAHX's effectiveness in diverse planning contexts.

\paragraph{RQ2: Representation Robustness.}

Our analysis indicates that PLAHX effectively mitigates context overflow and domain drift, as shown in Figure \ref{fig:rc}. In the Grippers domain, PLAHX achieves a high success rate with a minimum syntax (1.81\%) and semantic invalidity (22.54\%), in comparison to the syntax invalidity of 24.35\% of LLM-P-F(F/D). In the Rearrangement domain, PLAHX demonstrates robustness with an 87.85\% success rate and only 10.30\% semantic invalid plans, highlighting its resilience against representation inefficiencies.

\begin{table}[]
\setlength{\tabcolsep}{0.45mm}
\centering
\begin{tabular}{ccccc}
\toprule
                    & Blocks World & Tower of Hanoi & Grippers & Rearrangement              \\  \midrule
$|\mathcal{A}|$     &    22.6      &   42.36       &   17.93   & 20.89            \\  \midrule
$|\bar{A}^{(i)}|$   &    8.27       &   17.73      &   10.88   &  8.54       \\  \bottomrule
\end{tabular}
\caption{The average sizes of the action space $|\mathcal{A}|$ and the sub-space $|\bar{\mathcal{A}}^{(i)}|$ during the search of PLAHX for viable plans.}
\label{tab:space}
\end{table}

\begin{table}[]
\setlength{\tabcolsep}{0.45mm}
\centering
\begin{tabular}{ccccc}
\toprule
               & LLM-P-FF & LLM-P-FD & ISR-LLM & PLAHX     \\ \midrule
Blocks World   & 0.53   & 0.58  & -       & 0.42   \\ \midrule
Tower of Hanoi & 11.82  & 12.40 & -       & 11.50 \\ \midrule
Grippers       & 6.89   & 7.18  & -       & 46.29  \\ \midrule
Rearrangement     & 9.38   & 10.08 & 8.38    & 7.17   \\
\bottomrule
\end{tabular}
\caption{Comparison of CPU time costs (in seconds) for different planning methods across various domains. The results indicate that while PLAHX exhibits superior performance in Blocks World, Tower of Hanoi, and Rearrangement domains, it experiences a significant increase in computation time in the Grippers domain.}
\label{tab:time}
\end{table}

\paragraph{RQ3: Search Efficiency.}
We evaluate planning efficiency using three key metrics: average token cost (Table~\ref{tab:atc}), effective search space size (Table~\ref{tab:space}), and CPU runtime (Table~\ref{tab:time}). A detailed breakdown of token usage is provided in Appendix C.
As shown in Table~\ref{tab:atc}, PLAHX consistently achieves the lowest token consumption across all domains, with improvements ranging from 5.4\% to 98.8\% over the baselines. Table~\ref{tab:space} further demonstrates PLAHX’s ability to mitigate the exponential search bottleneck. For example, in the Blocks World domain, PLAHX reduces the effective search space to 8.27, compared to 22.6 for the best baseline. Correspondingly, Table~\ref{tab:time} shows that PLAHX achieves superior runtime efficiency in four domains. Overall, these results highlight significant contributions of PLAHX to improving both symbolic representation and planning efficiency, making it a robust solution for scalable task planning.



\subsubsection{Sensitivity Analysis and Ablation Studies}
Our studies reveal that the population size and the maximum subspace size are the pivotal parameters affecting the planning success and efficiency. The symbolic abstraction and the meta-heuristic planner within our approach enhance PDDL reasoning of LLMs and A* search performance, respectively, as detailed in Appendix D.

\section{Conclusion and Future Work}
In this work, we introduce PLAHX, a hybrid two-stage LLM-symbolic planning framework that combines compact symbolic abstraction with meta-heuristic subspace search to enable efficient and robust robotic task planning, particularly for long-horizon tasks. PLAHX directly addresses two core challenges in PDDL-based planning: (i) representation redundancy, which causes context overflow and domain drift in LLM pipelines, and (ii) the exponential search bottleneck inherent in classical symbolic planners. Extensive experiments across four standard robotic planning domains demonstrate that PLAHX consistently outperforms several strong baselines, achieving a significant increase in the average success rate, reduction of token consumption, and decrease in search complexity. 

However, PLAHX currently relies on predefined domain models and high-quality few-shot prompts. In the future, we aim to extend PLAHX to learning domain models from demonstrations or language \cite{DBLP:conf/icaps/OswaldSK00S24,PlanDark2025HuangLC25}, supporting probabilistic and partially observable planning settings, and integrating feedback loops for online adaptation and interactive refinement in dynamic environments, further enhancing the robustness and generality of LLM-symbolic planners in real-world applications.

\bibliography{ref}

\begin{thebibliography}{38}
\providecommand{\natexlab}[1]{#1}

\bibitem[{Allus and Unel(2025)}]{AStar2025Robotic}
Allus, A.; and Unel, M. 2025.
\newblock Angle-based multi-goal ordering and path-planning using an improved A-star algorithm.
\newblock \emph{Robotics Auton. Syst.}, 190: 105001.

\bibitem[{Brohan et~al.(2023)Brohan, Chebotar, Finn, Hausman, Herzog, Ho, Ibarz, Irpan, Jang, Julian et~al.}]{brohan2023can}
Brohan, A.; Chebotar, Y.; Finn, C.; Hausman, K.; Herzog, A.; Ho, D.; Ibarz, J.; Irpan, A.; Jang, E.; Julian, R.; et~al. 2023.
\newblock Do as i can, not as i say: Grounding language in robotic affordances.
\newblock In \emph{Conference on Robot Learning}, 287--318. PMLR.

\bibitem[{Celorrio et~al.(2012)Celorrio, de~la Rosa, Fern{\'{a}}ndez, Fern{\'{a}}ndez, and Borrajo}]{ReviewOfAutomatedPlanning}
Celorrio, S.~J.; de~la Rosa, T.; Fern{\'{a}}ndez, S.; Fern{\'{a}}ndez, F.; and Borrajo, D. 2012.
\newblock A review of machine learning for automated planning.
\newblock \emph{Knowl. Eng. Rev.}, 27(4): 433--467.

\bibitem[{Chevalier{-}Boisvert et~al.(2018)Chevalier{-}Boisvert, Bahdanau, Lahlou, Willems, Saharia, Nguyen, and Bengio}]{DBLP:journals/corr/abs-1810-08272}
Chevalier{-}Boisvert, M.; Bahdanau, D.; Lahlou, S.; Willems, L.; Saharia, C.; Nguyen, T.~H.; and Bengio, Y. 2018.
\newblock BabyAI: First Steps Towards Grounded Language Learning With a Human In the Loop.
\newblock \emph{CoRR}, abs/1810.08272.

\bibitem[{Chu et~al.(2025)Chu, Zhao, Weber, and Wermter}]{chu2025llm+map}
Chu, K.; Zhao, X.; Weber, C.; and Wermter, S. 2025.
\newblock LLM+MAP: Bimanual Robot Task Planning using Large Language Models and Planning Domain Definition Language.
\newblock \emph{arXiv preprint arXiv:2503.17309}.

\bibitem[{Corr{\^{e}}a, Pereira, and Seipp(2025)}]{LLMheuristic2025}
Corr{\^{e}}a, A.~B.; Pereira, A.~G.; and Seipp, J. 2025.
\newblock Classical Planning with LLM-Generated Heuristics: Challenging the State of the Art with Python Code.
\newblock \emph{CoRR}, abs/2503.18809.

\bibitem[{De~Filippo, Milano et~al.(2024)}]{de2024large}
De~Filippo, A.; Milano, M.; et~al. 2024.
\newblock Large language models for human-AI co-creation of robotic dance performances.
\newblock In \emph{IJCAI}, 7627--7635. International Joint Conferences on Artificial Intelligence.

\bibitem[{Fikes and Nilsson(1971)}]{DBLP:conf/ijcai/FikesN71}
Fikes, R.; and Nilsson, N.~J. 1971.
\newblock {STRIPS:} {A} New Approach to the Application of Theorem Proving to Problem Solving.
\newblock In Cooper, D.~C., ed., \emph{Proceedings of the 2nd International Joint Conference on Artificial Intelligence. London, UK, September 1-3, 1971}, 608--620. William Kaufmann.

\bibitem[{Fox and Long(2003)}]{PDDL2003}
Fox, M.; and Long, D. 2003.
\newblock {PDDL2.1:} An Extension to {PDDL} for Expressing Temporal Planning Domains.
\newblock \emph{J. Artif. Intell. Res.}, 20: 61--124.

\bibitem[{Garcia, Chen, and Schmid(2025)}]{garcia25gembench}
Garcia, R.; Chen, S.; and Schmid, C. 2025.
\newblock Towards Generalizable Vision-Language Robotic Manipulation: A Benchmark and LLM-guided 3D Policy.
\newblock In \emph{IEEE International Conference on Robotics and Automation (ICRA)}.

\bibitem[{Gerevini(2020)}]{IntroPDDL2020}
Gerevini, A.~E. 2020.
\newblock An Introduction to the Planning Domain Definition Language {(PDDL):} Book review.
\newblock \emph{Artif. Intell.}, 280: 103221.

\bibitem[{Glocker et~al.(2025)Glocker, H{\"o}nig, Hirschmanner, and Vincze}]{glocker2025llm}
Glocker, M.; H{\"o}nig, P.; Hirschmanner, M.; and Vincze, M. 2025.
\newblock Llm-empowered embodied agent for memory-augmented task planning in household robotics.
\newblock \emph{arXiv preprint arXiv:2504.21716}.

\bibitem[{Hart, Nilsson, and Raphael(1968)}]{hart1968formal}
Hart, P.~E.; Nilsson, N.~J.; and Raphael, B. 1968.
\newblock A formal basis for the heuristic determination of minimum cost paths.
\newblock \emph{IEEE transactions on Systems Science and Cybernetics}, 4(2): 100--107.

\bibitem[{Helmert(2006)}]{FastDownward2006}
Helmert, M. 2006.
\newblock The Fast Downward Planning System.
\newblock \emph{J. Artif. Intell. Res.}, 26: 191--246.

\bibitem[{Hoffmann(2001)}]{FastForward2001}
Hoffmann, J. 2001.
\newblock {FF:} The Fast-Forward Planning System.
\newblock \emph{{AI} Mag.}, 22(3): 57--62.

\bibitem[{Hoffmann and Nebel(2011)}]{FastForward2011}
Hoffmann, J.; and Nebel, B. 2011.
\newblock The {FF} Planning System: Fast Plan Generation Through Heuristic Search.
\newblock \emph{CoRR}, abs/1106.0675.

\bibitem[{Howey, Long, and Fox(2004)}]{VAL2004}
Howey, R.; Long, D.; and Fox, M. 2004.
\newblock {VAL:} Automatic Plan Validation, Continuous Effects and Mixed Initiative Planning Using {PDDL}.
\newblock In \emph{16th {IEEE} International Conference on Tools with Artificial Intelligence {(ICTAI} 2004), 15-17 November 2004, Boca Raton, FL, {USA}}, 294--301. {IEEE} Computer Society.

\bibitem[{Huang, Lipovetzky, and Cohn(2025)}]{PlanDark2025HuangLC25}
Huang, S.; Lipovetzky, N.; and Cohn, T. 2025.
\newblock Planning in the Dark: LLM-Symbolic Planning Pipeline Without Experts.
\newblock In Walsh, T.; Shah, J.; and Kolter, Z., eds., \emph{AAAI-25, Sponsored by the Association for the Advancement of Artificial Intelligence, February 25 - March 4, 2025, Philadelphia, PA, {USA}}, 26542--26550. {AAAI} Press.

\bibitem[{Huang et~al.(2022)Huang, Abbeel, Pathak, and Mordatch}]{huang2022language}
Huang, W.; Abbeel, P.; Pathak, D.; and Mordatch, I. 2022.
\newblock Language models as zero-shot planners: Extracting actionable knowledge for embodied agents.
\newblock In \emph{International conference on machine learning}, 9118--9147. PMLR.

\bibitem[{Kahneman(2011)}]{kahneman2011thinking}
Kahneman, D. 2011.
\newblock \emph{Thinking, fast and slow}.
\newblock macmillan.

\bibitem[{Kambhampati et~al.(2024)Kambhampati, Valmeekam, Guan, Verma, Stechly, Bhambri, Saldyt, and Murthy}]{kambhampati2024position}
Kambhampati, S.; Valmeekam, K.; Guan, L.; Verma, M.; Stechly, K.; Bhambri, S.; Saldyt, L.~P.; and Murthy, A.~B. 2024.
\newblock Position: LLMs can’t plan, but can help planning in LLM-modulo frameworks.
\newblock In \emph{Forty-first International Conference on Machine Learning}.

\bibitem[{Kannan, Venkatesh, and Min(2024)}]{kannan2024smart}
Kannan, S.~S.; Venkatesh, V.~L.; and Min, B.-C. 2024.
\newblock Smart-llm: Smart multi-agent robot task planning using large language models.
\newblock In \emph{2024 IEEE/RSJ International Conference on Intelligent Robots and Systems (IROS)}, 12140--12147. IEEE.

\bibitem[{Latif(2024)}]{latif20243p}
Latif, E. 2024.
\newblock 3p-llm: Probabilistic path planning using large language model for autonomous robot navigation.
\newblock \emph{arXiv preprint arXiv:2403.18778}.

\bibitem[{Liu et~al.(2023)Liu, Jiang, Zhang, Liu, Zhang, Biswas, and Stone}]{liu2023llm+p}
Liu, B.; Jiang, Y.; Zhang, X.; Liu, Q.; Zhang, S.; Biswas, J.; and Stone, P. 2023.
\newblock LLM+P: Empowering large language models with optimal planning proficiency.
\newblock \emph{arXiv preprint arXiv:2304.11477}.

\bibitem[{Mahdavi et~al.(2024)Mahdavi, Aoki, Tang, and Cao}]{DomainDrift2024}
Mahdavi, S.; Aoki, R.; Tang, K.; and Cao, Y. 2024.
\newblock Leveraging Environment Interaction for Automated {PDDL} Generation and Planning with Large Language Models.
\newblock \emph{CoRR}, abs/2407.12979.

\bibitem[{McDermott(2000)}]{DBLP:journals/aim/McDermott00}
McDermott, D.~V. 2000.
\newblock The 1998 {AI} Planning Systems Competition.
\newblock \emph{{AI} Mag.}, 21(2): 35--55.

\bibitem[{Oswald et~al.(2024)Oswald, Srinivas, Kokel, Lee, Katz, and Sohrabi}]{DBLP:conf/icaps/OswaldSK00S24}
Oswald, J.~T.; Srinivas, K.; Kokel, H.; Lee, J.; Katz, M.; and Sohrabi, S. 2024.
\newblock Large Language Models as Planning Domain Generators.
\newblock In Bernardini, S.; and Muise, C., eds., \emph{Proceedings of the Thirty-Fourth International Conference on Automated Planning and Scheduling, {ICAPS} 2024, Banff, Alberta, Canada, June 1-6, 2024}, 423--431. {AAAI} Press.

\bibitem[{Segovia-Aguas, Jiménez, and Jonsson(2021)}]{Segovia-Aguas_Jiménez_Jonsson_2021}
Segovia-Aguas, J.; Jiménez, S.; and Jonsson, A. 2021.
\newblock Generalized Planning as Heuristic Search.
\newblock \emph{Proceedings of the International Conference on Automated Planning and Scheduling}, 31(1): 569--577.

\bibitem[{Silver and Chitnis(2020)}]{silver2020pddlgym}
Silver, T.; and Chitnis, R. 2020.
\newblock PDDLGym: Gym Environments from PDDL Problems.
\newblock In \emph{International Conference on Automated Planning and Scheduling (ICAPS) PRL Workshop}.

\bibitem[{Silver et~al.(2022)Silver, Hariprasad, Shuttleworth, Kumar, Lozano-P{\'e}rez, and Kaelbling}]{silver2022pddl}
Silver, T.; Hariprasad, V.; Shuttleworth, R.~S.; Kumar, N.; Lozano-P{\'e}rez, T.; and Kaelbling, L.~P. 2022.
\newblock PDDL planning with pretrained large language models.
\newblock In \emph{NeurIPS 2022 foundation models for decision making workshop}.

\bibitem[{Sloman(1996)}]{sloman1996empirical}
Sloman, S.~A. 1996.
\newblock The empirical case for two systems of reasoning.
\newblock \emph{Psychological bulletin}, 119(1): 3.

\bibitem[{Song et~al.(2023)Song, Wu, Washington, Sadler, Chao, and Su}]{Song_2023_ICCV}
Song, C.~H.; Wu, J.; Washington, C.; Sadler, B.~M.; Chao, W.-L.; and Su, Y. 2023.
\newblock LLM-Planner: Few-Shot Grounded Planning for Embodied Agents with Large Language Models.
\newblock In \emph{Proceedings of the IEEE/CVF International Conference on Computer Vision (ICCV)}, 2998--3009.

\bibitem[{Tantakoun, Zhu, and Muise(2025)}]{DBLP:journals/corr/abs-2503-18971}
Tantakoun, M.; Zhu, X.; and Muise, C. 2025.
\newblock LLMs as Planning Modelers: {A} Survey for Leveraging Large Language Models to Construct Automated Planning Models.
\newblock \emph{CoRR}, abs/2503.18971.

\bibitem[{Touvron et~al.(2023)Touvron, Lavril, Izacard, Martinet, Lachaux, Lacroix, Rozi{\`{e}}re, Goyal, Hambro, Azhar, Rodriguez, Joulin, Grave, and Lample}]{LLaMA2023}
Touvron, H.; Lavril, T.; Izacard, G.; Martinet, X.; Lachaux, M.; Lacroix, T.; Rozi{\`{e}}re, B.; Goyal, N.; Hambro, E.; Azhar, F.; Rodriguez, A.; Joulin, A.; Grave, E.; and Lample, G. 2023.
\newblock LLaMA: Open and Efficient Foundation Language Models.
\newblock \emph{CoRR}, abs/2302.13971.

\bibitem[{Valmeekam et~al.(2023)Valmeekam, Marquez, Sreedharan, and Kambhampati}]{valmeekam2023planning}
Valmeekam, K.; Marquez, M.; Sreedharan, S.; and Kambhampati, S. 2023.
\newblock On the planning abilities of large language models-a critical investigation.
\newblock \emph{Advances in Neural Information Processing Systems}, 36: 75993--76005.

\bibitem[{Valmeekam et~al.(2022)Valmeekam, Olmo, Sreedharan, and Kambhampati}]{valmeekam2022large}
Valmeekam, K.; Olmo, A.; Sreedharan, S.; and Kambhampati, S. 2022.
\newblock Large language models still can't plan (a benchmark for LLMs on planning and reasoning about change).
\newblock In \emph{NeurIPS 2022 Foundation Models for Decision Making Workshop}.

\bibitem[{Zeng et~al.(2020)Zeng, Florence, Tompson, Welker, Chien, Attarian, Armstrong, Krasin, Duong, Sindhwani, and Lee}]{DBLP:conf/corl/ZengFTWCAAKDSL20}
Zeng, A.; Florence, P.; Tompson, J.; Welker, S.; Chien, J.; Attarian, M.; Armstrong, T.; Krasin, I.; Duong, D.; Sindhwani, V.; and Lee, J. 2020.
\newblock Transporter Networks: Rearranging the Visual World for Robotic Manipulation.
\newblock In Kober, J.; Ramos, F.; and Tomlin, C.~J., eds., \emph{4th Conference on Robot Learning, CoRL 2020, 16-18 November 2020, Virtual Event / Cambridge, MA, {USA}}, volume 155 of \emph{Proceedings of Machine Learning Research}, 726--747. {PMLR}.

\bibitem[{Zhou et~al.(2024)Zhou, Song, Yao, Shu, and Ma}]{zhou2024isr}
Zhou, Z.; Song, J.; Yao, K.; Shu, Z.; and Ma, L. 2024.
\newblock ISR-LLM: Iterative self-refined large language model for long-horizon sequential task planning.
\newblock In \emph{2024 IEEE International Conference on Robotics and Automation (ICRA)}, 2081--2088. IEEE.

\end{thebibliography}

\clearpage

\appendix

\section*{Appendix}

\section{Benchmark Details}
In this work, we construct a set of benchmarks to test the planning performance of LLM-symbolic planers given the open-ended natural language instructions.
The benchmark covers four classic planning domains, detailed as follows:
\begin{itemize}
    \item Blocks: Given a set of piles of colored blocks on a table, a robot is tasked with stacking them into a specified target configuration.
    \item Hanoi: The target of this problem is to move a stack of disks from one rod to another, following specific rules.
    \item Grippers: A robot with two grippers is given a task to operate objects among different rooms.
    \item Rearrangement: This is a variant of Blocks, where the robot must arrange blocks into bowls to achieve a specified target configuration.
\end{itemize}

Table \ref{tab-benchmark} provide details of the dataset where the PDDL data is constructed based on \cite{silver2020pddlgym} and the natural lanague of task descriptions are generated based on \cite{DBLP:conf/corl/ZengFTWCAAKDSL20,DBLP:journals/corr/abs-1810-08272,garcia25gembench}. For each domain, we automatically generate $m$ diverse task instances. Each instance includes (i) a natural language instruction, (ii) a corresponding PDDL problem specification, and (iii) the requisite domain model. The example of PDDL domain and problem files are shown in Figure \ref{fig:pddl1} and \ref{fig:pddl2}.

\setcounter{figure}{3}
\begin{figure}[!ht]
    \centering
    \captionsetup[subfloat]{labelformat=parens} 
    \subfloat[Blocks]{
        \includegraphics[width=0.9\linewidth]{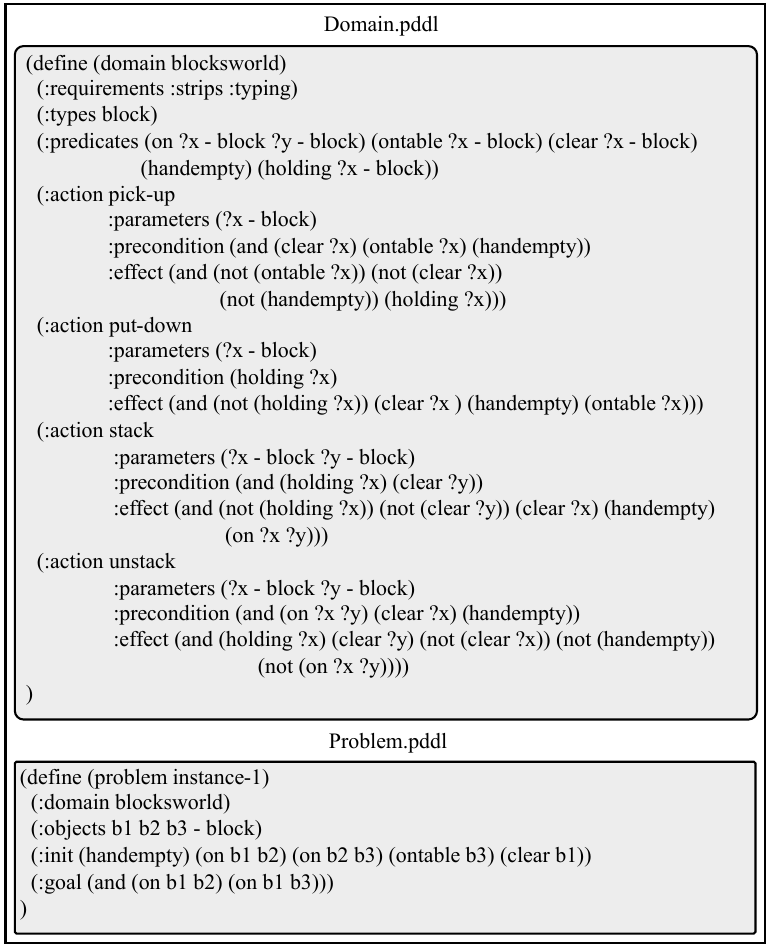}
        \label{fig:blocksworld}
    }\\
    \subfloat[Hanoi]{
        \includegraphics[width=0.9\linewidth]{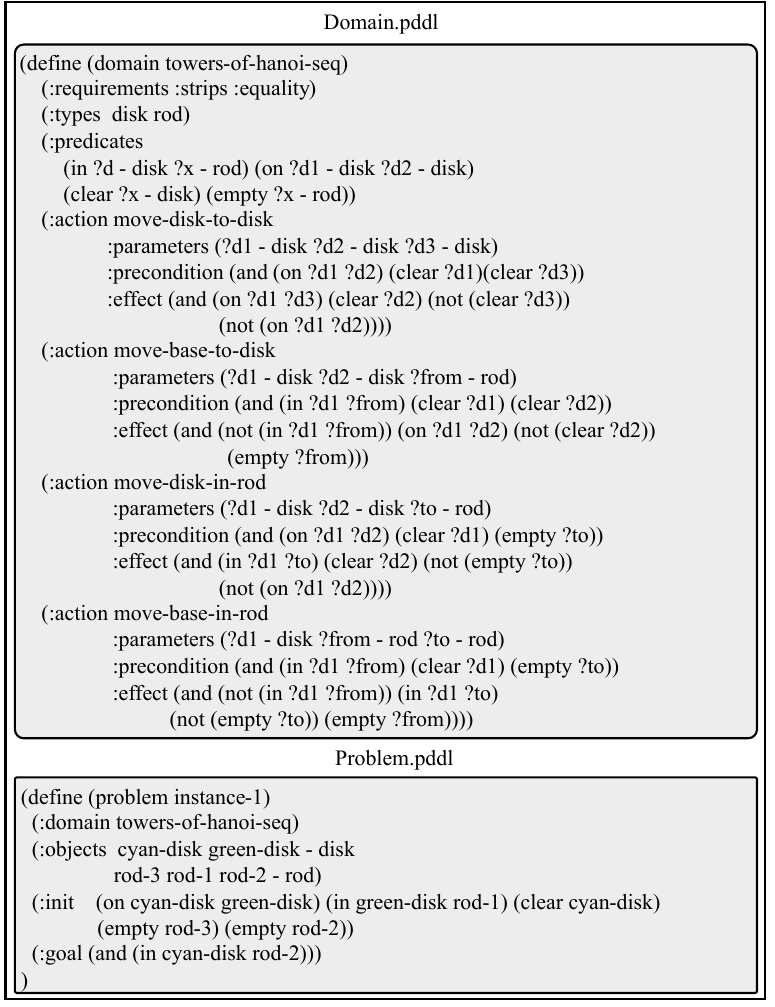}
        \label{fig:Hanoi}
    }
    \caption{The PDDL model. (a) Blocks. (b) Hanoi.}
    \label{fig:pddl1}
\end{figure}

\begin{figure}[!ht]
    \centering
    \captionsetup[subfloat]{labelformat=parens} 
    \subfloat[Grippers]{
        \includegraphics[width=0.9\linewidth]{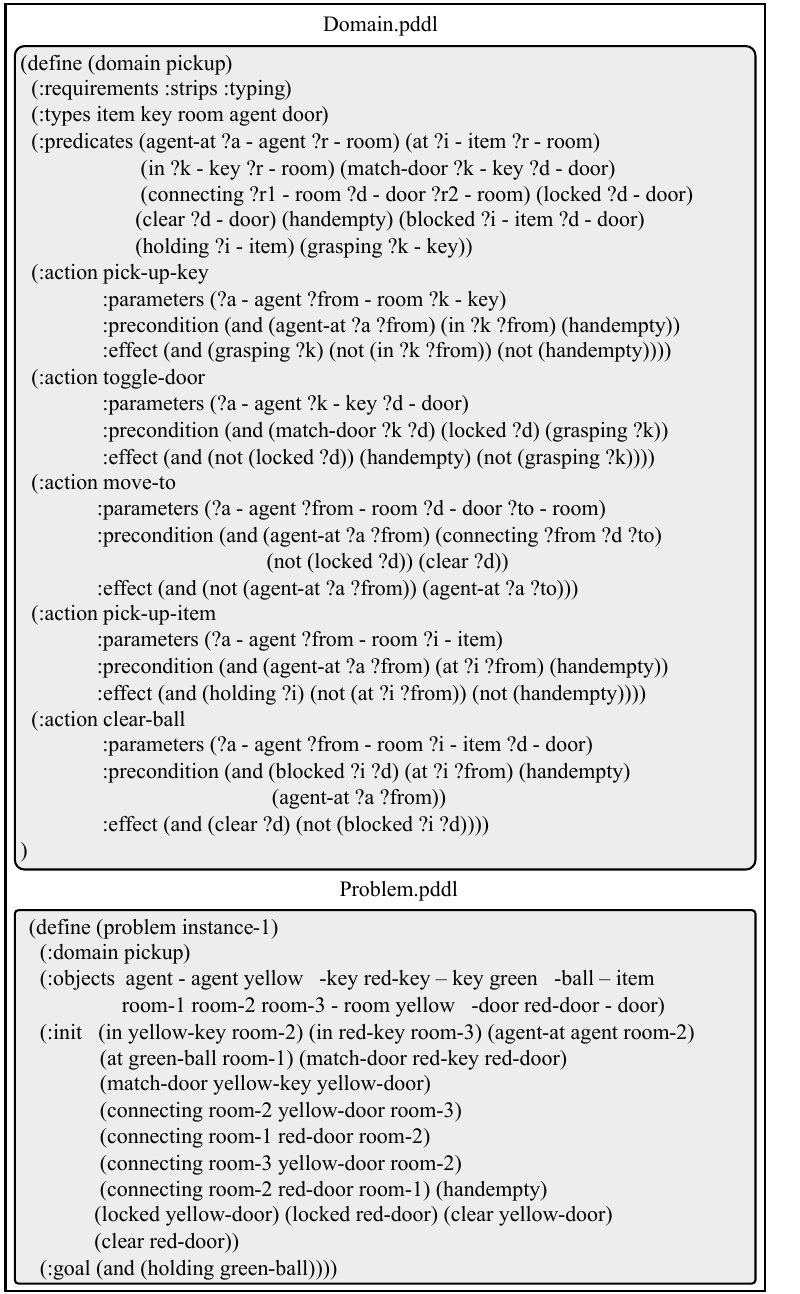}
        \label{fig:grippers}
    }\\
    \subfloat[Reaarangement]{
        \includegraphics[width=0.9\linewidth]{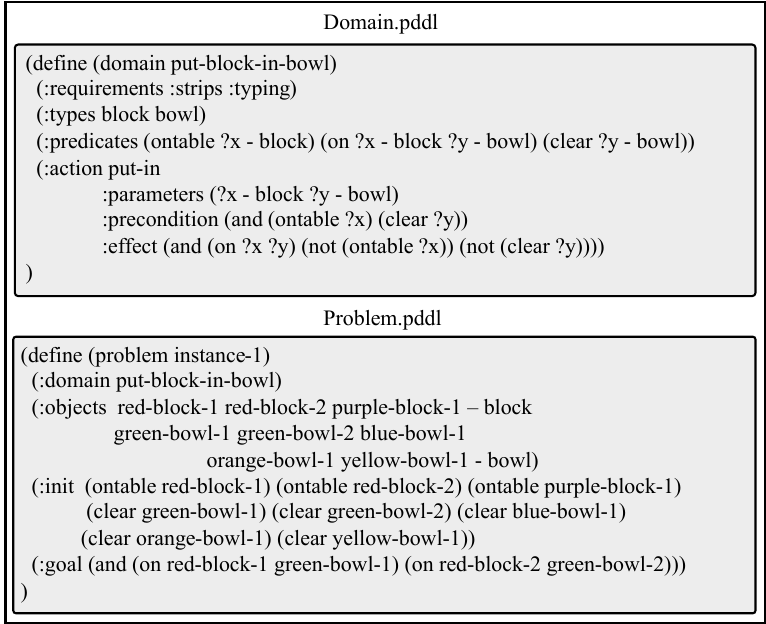}
        \label{fig:rearrangement}
    }
    \caption{The PDDL model. (a) Grippers. (b) Rearrangement.}
    \label{fig:pddl2}
\end{figure}

\section{Experimental Setup}
We conduct on LLaMA model (\texttt{Llama-3-8B}) \cite{LLaMA2023} with temperature $0.0$, and set max tokens $5000$, as the context window limit. For In-context learning, we provide 6 examples to the model via few-shot prompting. Experiments are run on a 24-core Intel i7-13700KF (3.2–5.4 GHz) and the LLM is deployed on an NVIDIA H20 (96 GB). 
Additionally, the parameters involved include the population size, the maximum length of subspaces, the maximum number of iterations, mutation parameter, penalty magnitude, decay rate, and lower bound. Detailed settings of all parameters are shown in Table~\ref{tab-parameter}.

\setcounter{table}{4}
\begin{table}[]
\setlength{\tabcolsep}{0.5mm}
\centering
\begin{tabular}{c c c }
\toprule
\textbf{Parameter} & \textbf{Value} & \textbf{Description}\\ 
\midrule
${temperature}$ & 0.0  & \makecell{The value of\\temperature to LLMs.} \\ \midrule
$\ell_{\max}$ & 5000  & \makecell{The maximum context\\length in tokens of LLMs.} \\
\midrule
$N_{pop}$ & 20 & \makecell{The size of population.}\\ \midrule
$l_{max}$ & $\left\lfloor |\mathcal{A}|/2\right\rfloor$ & \makecell{The maximum length of\\ subspaces.}\\ \midrule
$M$ & 100 & \makecell{The maximum of\\the iteration.}\\ \midrule
$N_{stop}$ & 2 & \makecell{The threshold for algorithm\\ stopping early.}\\ \midrule
$p_{\text{mut}}$ & \makecell{$0.05+\frac{M-m}{10*M}$} & \makecell{The calculation of mutation\\ probability where $m$ is the\\ iteration number.}\\ \midrule
$\rho$ & $0.1\times e^{-0.025\cdot m}$ & \makecell{The penalty magnitude.}\\
$\tau$ & 200 & \makecell{The decay rate.}\\ \midrule
$\ell_b$ & 0.1 & \makecell{The lower bound.}\\
\bottomrule
\end{tabular}
\caption{The settings of all parameters involved in experiments.}
\label{tab-parameter}
\end{table}

\begin{table*}[]
\setlength{\tabcolsep}{1mm}
\centering
\begin{tabular}{c p{3.cm} p{7.3cm} c c c}
\toprule
\textbf{Domain} & \textbf{Example Goal} & \textbf{Example Initial Observation} & \textbf{Plan Length} & $|S|$ & $m$\\ 
\midrule
Blocks             & Stack 2 rose blocks. & I have 2 rose blocks (b1, b2) and other blocks. Block b1 is on the table. Block b2 is on the table.                                                             &    $2\sim 12$         &  $23\sim69$ & 299 \\ 
\midrule
Hanoi           & Move the gray disk in rod 3.                                & Gray disk on top of blue disk. blue disk in rod 1. The disks can be moved in rod 1, rod 2, rod 3.                                                                            &     $1\sim 3$        & $65\sim132$  & 757\\
\midrule
Grippers                  & Pick up the purple box.                                           & Room 1 has grey key, agent. Room 2 has purple box. The door connecting Room 1 and Room 2 is locked. &   $4\sim8$          & $28\sim81$ & 896\\
\midrule
Allocation               & Put the blue blocks in gray bowls.                          & There is a gray bowl 1, gray bowl 2, blue block 1, blue block 2, red block 1, green bowl 1, orange bowl 1, yellow block 1.                                                   &    $2\sim4$        & $32\sim84$ & 497\\ 
\bottomrule
\end{tabular}
\caption{Overview of planning tasks across various environments, including example goals, initial observations, the average state space size $|S|$, and the number of task instances $m$.}
\label{tab-benchmark}
\end{table*}

\section{Analysis of Symbolic Representation}
In this section, we delve into the analysis of how token consumption and translation error classes manifest differently under zero-shot and few-shot prompting scenarios. Our investigation focuses on the impact of varying the number of task exemplars (1 through 6) within a larger context length of 8000 tokens in LLMs. We have omitted chain-of-thought prompting from our analysis due to its tendency to exceed the token limit with extended outputs.

\begin{figure*}[!ht]
    \centering
    \captionsetup[subfloat]{labelformat=parens} 
    \subfloat[Blocks]{
        \includegraphics[width=1\linewidth]{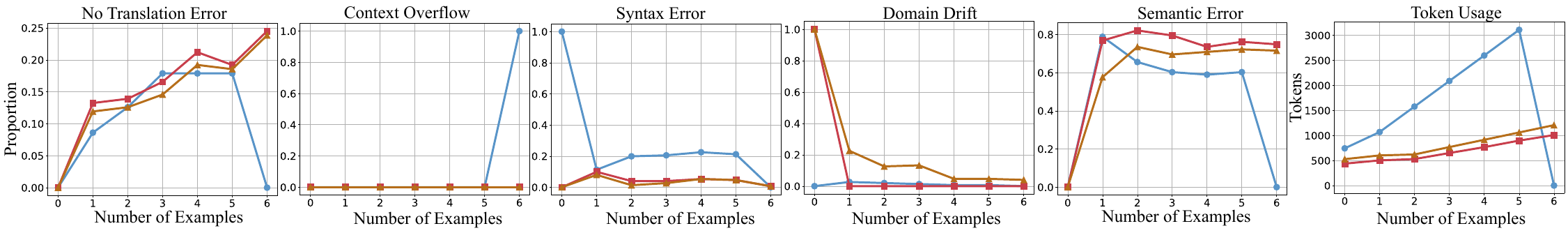}
        \label{fig:blocks}
    }\\
    \subfloat[Hanoi]{
        \includegraphics[width=1\linewidth]{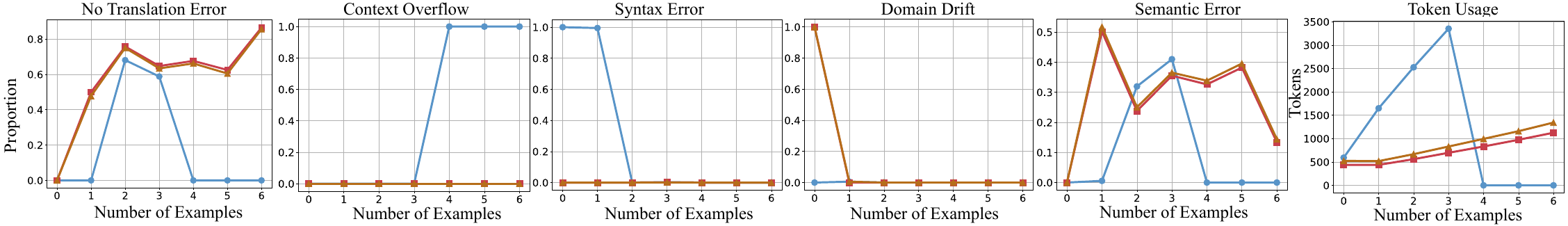}
        \label{fig:hanoi}
    }\\
    \subfloat[Grippers]{
        \includegraphics[width=1\linewidth]{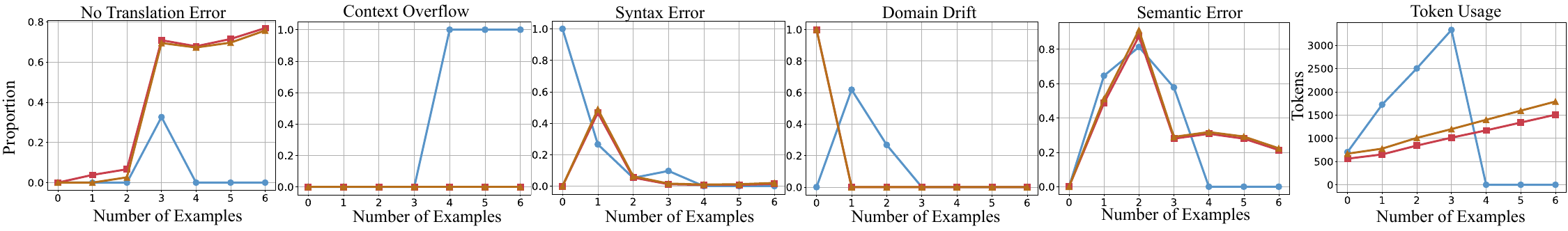}
        \label{fig:Grippers}
    }\\
    \subfloat[Rearrangement]{
        \includegraphics[width=1\linewidth]{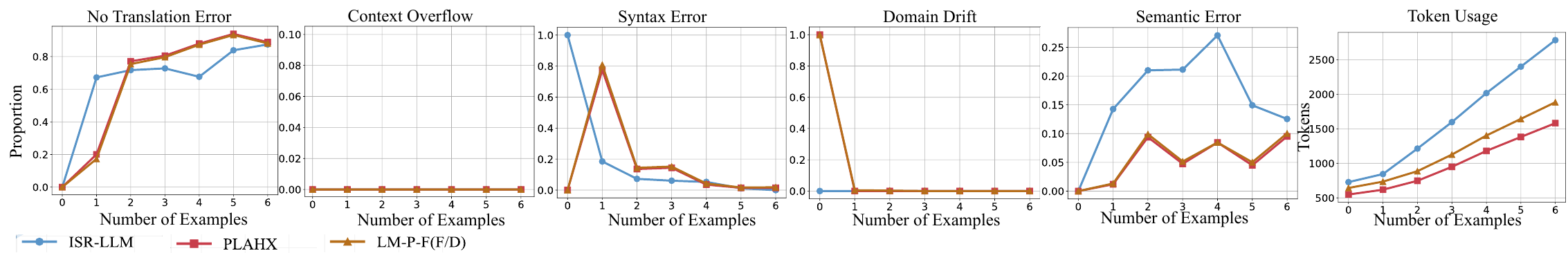}
        \label{fig:Rearrangement}
    }
    \caption{The classes of the result (proportion) and token usages across domains.}
    \label{fig:token_usage}
\end{figure*}

The results, as illustrated in Figure \ref{fig:token_usage}, provide a comprehensive overview of the performance metrics across different domains: Blocks, Hanoi, Grippers, and Rearrangement. The analysis highlights key trends in token usage and the proportion of classes of translation errors within the token horizon, offering insights into the effectiveness of different prompting strategies:
\begin{itemize}
    \item (Token Usage and Task Performance)
    A consistent trend across all domains is the increase in token usage with the addition of more task exemplars. This pattern is particularly evident in the ISR-LLM and LLM-P-F(F/D), where the token consumption rises significantly as the number of examples increases. This increase in token usage is directly correlated with the ability to solve tasks within the allowed horizon, suggesting that a higher number of exemplars provides the model with a richer context, thereby enhancing its problem-solving capabilities.
    \item (Translation Errors and Exemplar Influence)
    The impact of task exemplars on translation errors is a critical aspect of our analysis. In the Blocks domain, both ISR-LLM and PLAIX exhibit a decrease in semantic errors as the number of examples increases, indicating an improved understanding of task semantics. This trend is mirrored in the Hanoi and Grippers domains, where the reduction in semantic errors with increased exemplars underscores the importance of exemplar-based learning in enhancing model performance.    
\end{itemize}
Our study highlights the importance of symbolic representations in how LLMs use tokens. PLAHX excels in managing token use and minimizing semantic errors among few-shot strategies, which is vital for tasks needing a balance of computational resources and complexity. The strength of PLAHX comes from using more examples to give the model richer context, enhancing problem-solving within limits.

\section{Ablation Study}

\subsection{Component Analysis}
In this section, we analyze the effectiveness of the two core components of the proposed PLAHX: (1) the concise \mbox{symbolic} representation generated during the translation stage, and (2) the meta-heuristic search employed in the planning stage. Instead of conducting ablation studies by removing each component, we leverage the results from our previous experiments to assess their individual contributions to the overall performance:

\begin{itemize}
    \item {(Concise Symbolic Representation):}
    The effectiveness of the concise symbolic representation generated during the translation stage can be inferred from the performance comparison between PLAHX and LLM-P-F(F/D) as depicted in Figure \ref{fig:token_usage}. PLAHX consistently demonstrates a lower token usage but a higher translation success across various domains, including Blocks, Hanoi, Grippers, and Rearrangement. This suggests that the symbolic representation produced by PLAHX is more precise and efficient, leading to fewer semantic discrepancies and better task execution within the allowed token horizon.

    \item {(Meta-Heuristic Search):}
    The efficiency of the meta-heuristic search employed in the planning stage is evidenced by the analysis of action space sizes and CPU time costs as shown in Tables 3 and 4. PLAHX exhibits a significantly smaller average size of the action space $|\mathcal{A}|$ and the sub-space $|\bar{\mathcal{A}}^{(i)}|$ across domains like Blocks, Hanoi, and Rearrangement, indicating a more focused and efficient search strategy. Additionally, the CPU time costs in Table 4 reveal that PLAHX performs comparably or better than other methods in terms of computation time, except in the Grippers domain where it experiences a notable increase. This suggests that the meta-heuristic search in PLAHX is generally effective in optimizing planning efficiency, although further refinement may be needed for certain complex tasks.
\end{itemize}

\begin{figure*}[!ht]
    \centering
    \captionsetup[subfloat]{labelformat=parens} 
    \subfloat[Planning success (\%)]{
        \includegraphics[width=0.4\linewidth]{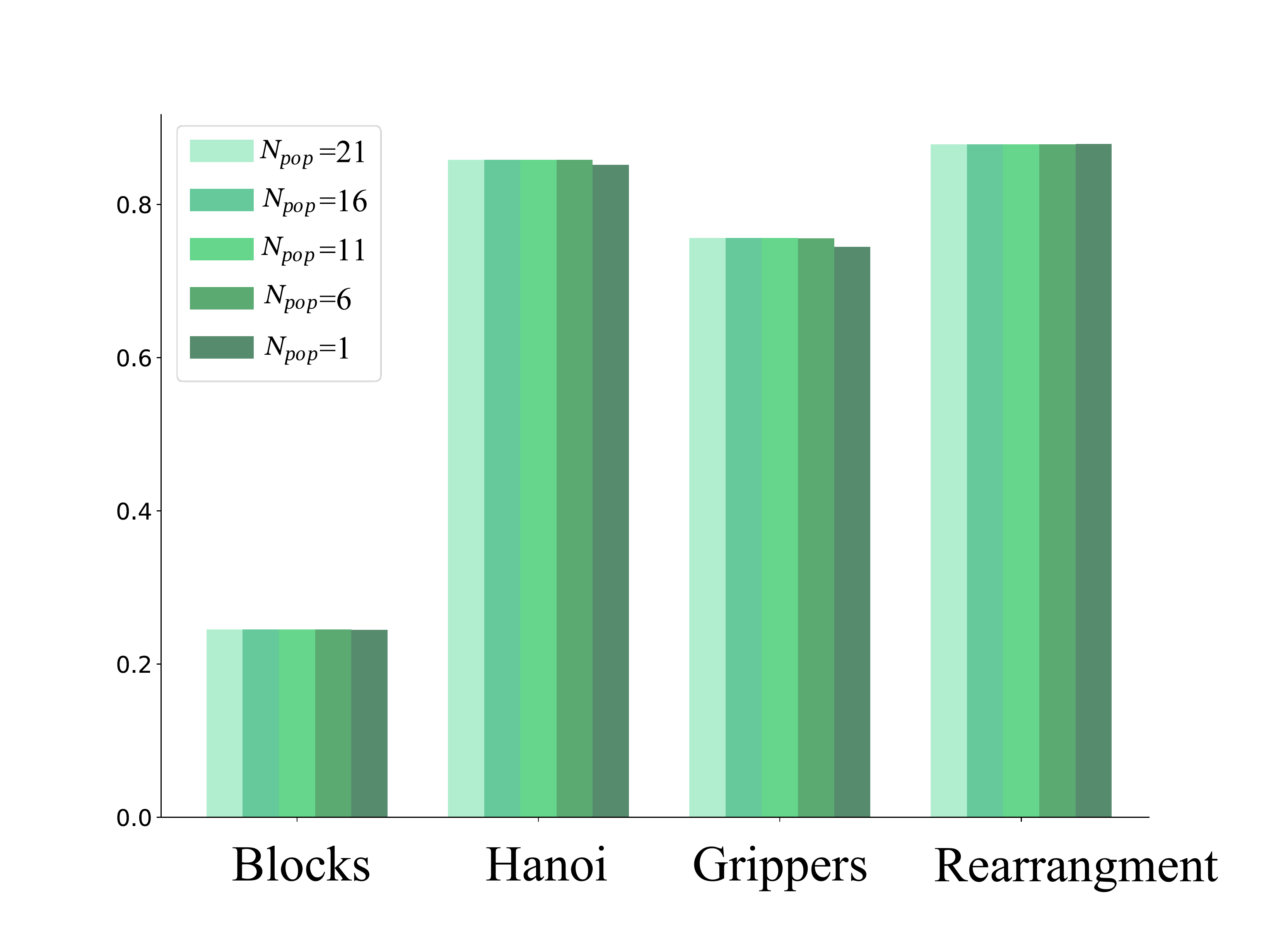}
        \label{fig:popsize-sr}
    }
    \subfloat[Planning time (in seconds)]{
        \includegraphics[width=0.4\linewidth]{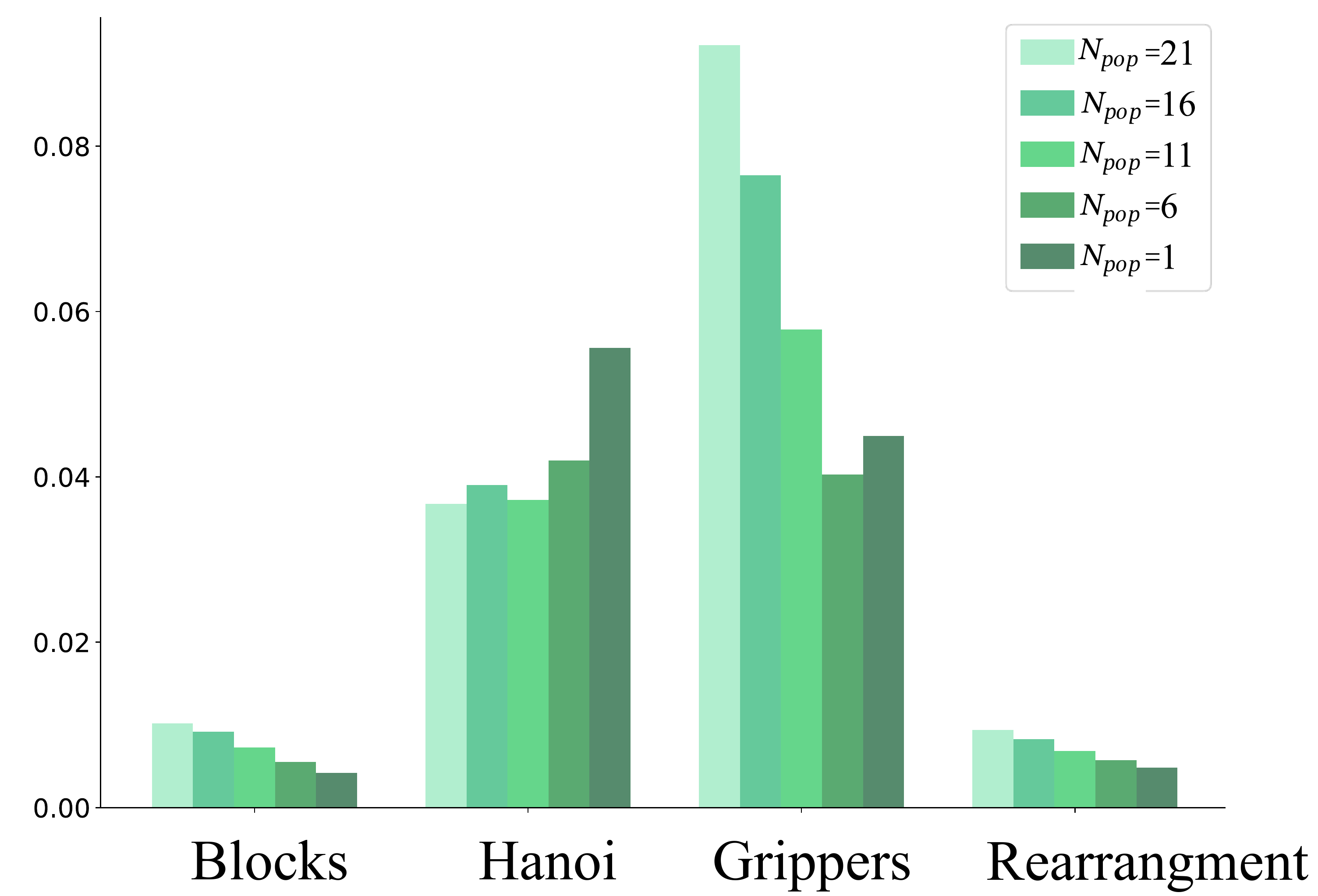}
        \label{fig:popsize-time}
    }
    \caption{The planning success rate and time (on average) of different population sizes $N_{pop}$.}
    \label{fig:ablation-popsize}
\end{figure*}

\begin{figure*}[!ht]
    \centering
    \captionsetup[subfloat]{labelformat=parens} 
    \subfloat[Planning success (\%)]{
        \includegraphics[width=0.35\linewidth]{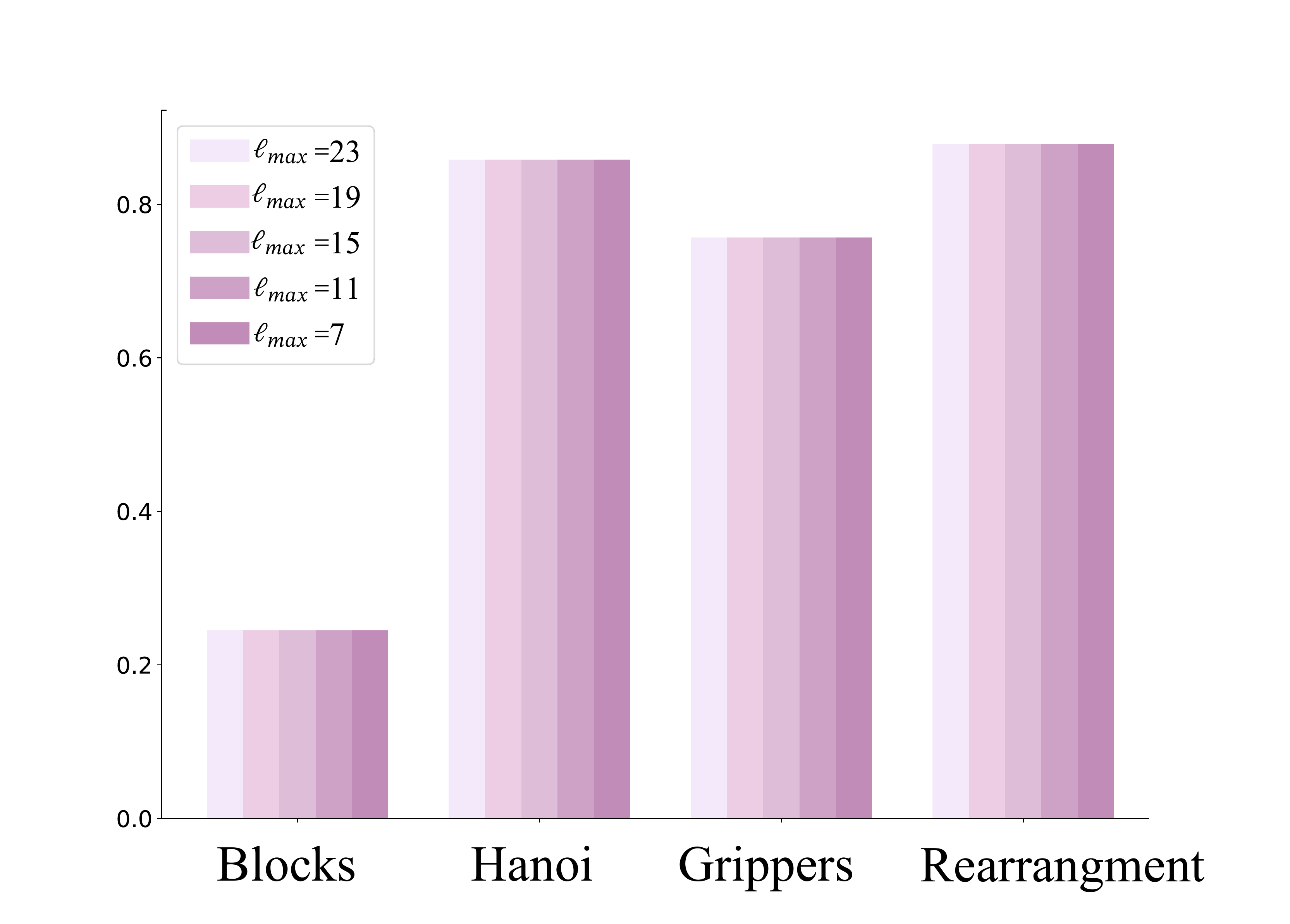}
        \label{fig:subsize-sr}
    }
    \subfloat[Planning time (in seconds)]{
        \includegraphics[width=0.35\linewidth]{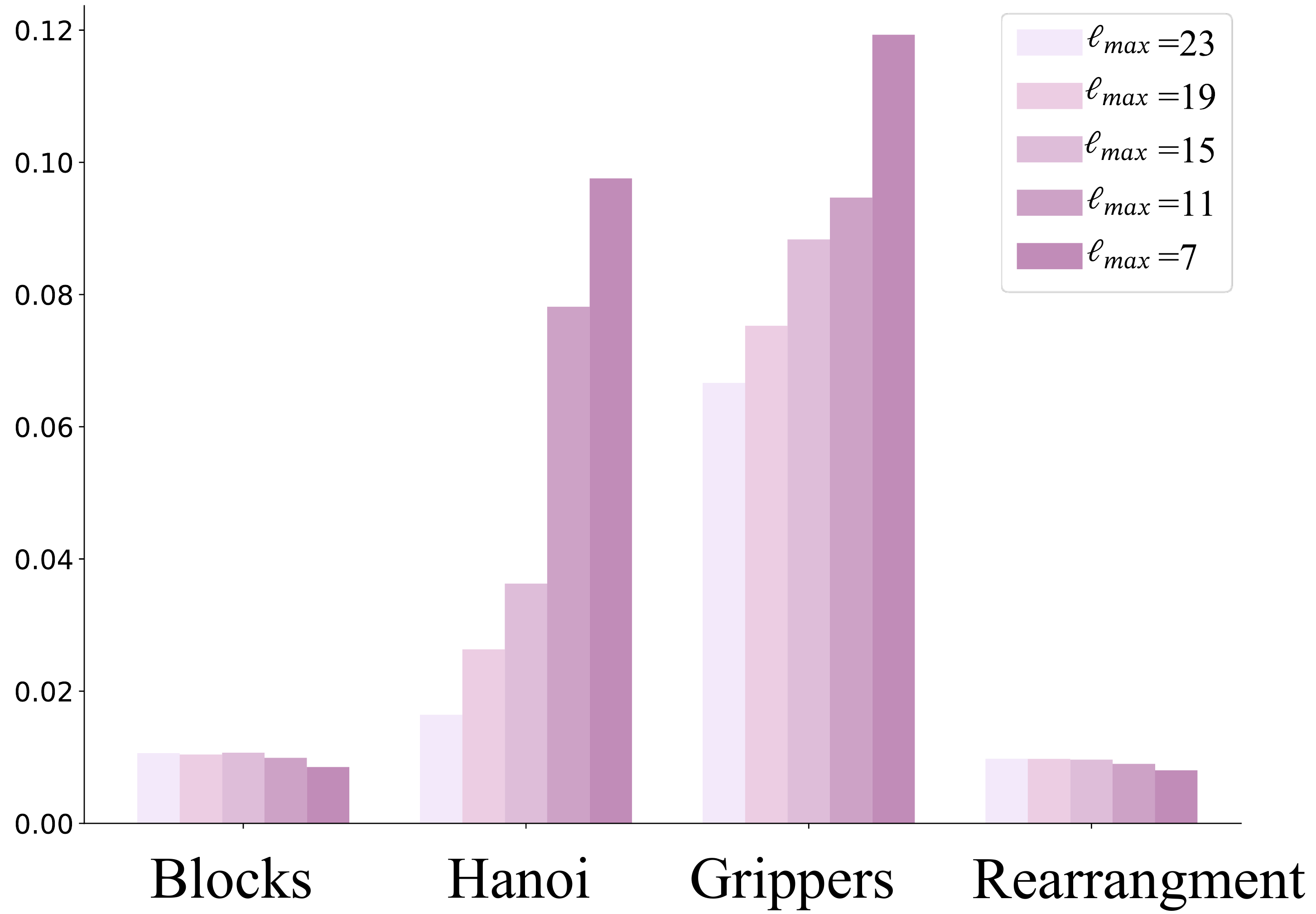}
        \label{fig:subsize-time}
    }
    \caption{The planning success rate and time (on average) of different maximum sizes of subspaces $\ell_{max}$.}
    \label{fig:ablation-subsize}
\end{figure*}

\subsection{Parameter Sensitivity Analysis}
We vary the population size ($N_{pop}$) and the subspace size ($\ell_{max}$) of the meta-heuristic search to assess robustness. For each ($N_{pop},\ell_{max}$) configuration we record success rate and running time, identifying the ranges within which performance is stable and the points where degradation occurs.

As depicted in Figure~\ref{fig:ablation-popsize}, in terms of success rate, the planner is remarkably insensitive to population size $N_{pop}$ within the tested interval $[1, 21]$. Runtime behaviour, however, diverges by task structure. In Blocks, Grippers and Rearrangement, where the search space is dominated by many short, interchangeable actions (pick, place, move), increasing $N_{pop}$ introduces only a mild, almost linear growth in running time. The additional individuals primarily generate redundant plans without enlarging the effective frontier. Conversely, in Hanoi, where actions are strictly sequential and each decision has long-range consequences, a larger population provides more diverse high-quality partial plans early on, pruning the state space aggressively. Consequently, runtime decreases in an almost linear fashion as $N_{pop}$ grows, dropping by roughly 40\% from $N_{pop}=1$ to $N_{pop}$.

Performance degrades only at the extreme $N_{pop}=1$: success rates remain stable across all domains, yet runtimes double in Blocks, Grippers and Rearrangement, while Hanoi’s runtime rises sharply owing to the loss of beneficial diversity. Beyond $N_{pop}=6$ the curves flatten, indicating that further population growth yields negligible improvement in any domain.

Figure~\ref{fig:ablation-subsize} shows that subspace size $\ell_{max}$ has virtually no influence on success rate; all domains remain within a 2\% band across the entire tested range $\ell_{max} \in [7, 23]$. However, runtime responds markedly. In Blocks and Rearrangement, where actions are short and local (pick/place in Blocks, put-into-bowl in Rearrangement), a smaller latent subspace is sufficient to encode the limited action dependencies.
Consequently, runtime decreases almost linearly as $\ell_{max}$ shrinks from 23 to 7, because the planner evaluates fewer, yet still adequate, latent operators at each step. By contrast, Hanoi and Grippers feature long-horizon dependencies (disk order constraints in Hanoi and room-connectivity constraints in Grippers) that demand richer latent representations. Reducing $\ell_{max}$ forces the planner to reconstruct these dependencies from a compressed subspace, incurring additional decoding overhead. Runtime therefore rises linearly as $\ell_{max}$ decreases, with the sharpest increase observed in Hanoi once $\ell_{max}$ falls below 11.

Only when $\ell_{max}\le 5$, the compression become too aggressive for domains, causing a sudden spike in time that signals the lower bound of the usable range.

\paragraph{Impact of Parallel Subspace Search on the Exponential Bottleneck}
The empirical trends in Figures~\ref{fig:ablation-popsize} and~\ref{fig:ablation-subsize} can be re-interpreted through the lens of Parallel Subspace Compression Ratio ($PSCR$, Definition~4). Across all four domains, the planner keeps the success rate almost constant while exhibiting distinct runtime. This is precisely the signature of a successful parallel-subspace compression whose magnitude is governed by the domain-specific ratio $PSCR(D)$.

\setcounter{definition}{3}
\begin{definition}
  The effect of relieving exponential bottleneck through parallel subspace search is defined as Parallel Subspace Compression Ratio (PSCR):
  \begin{equation}
    PSCR(D) = \frac{|\mathcal{A}|}{N_{pop} \cdot \max |\mathcal{A}^j|}, \quad j \in \{1,\ldots N_{pop}\}.
    \label{eq:pscr}
  \end{equation}
  $PSCR(D) > 1$ denotes that the partition and parallel strategy compresses the search scale to an exponential fraction of the original space, thereby reducing the heuristic search of each subspace from $T_{\text{search}}=\Omega\!\bigl(|\mathcal{A}|^{1-\gamma}\bigr)$ to $T_{\text{sub}}=\Omega\!\bigl(\frac{|\mathcal{A}|}{N_{pop}}^{1-\gamma}\bigr)$. The overall running time changes from exponential to approximately linear acceleration (ideally close to $N_{pop}$ times).
\end{definition}

\paragraph{Population size $N_{pop}$:}
Each individual explores an independent latent subspace $\mathcal{A}^{(j)}$.
\begin{itemize}
    \item In Blocks, Grippers, and Rearrangement, actions are interchangeable and the branching factor is high, so the original search space $|\mathcal{A}|$ is large. Here $PSCR(D) \approx \frac{|\mathcal{A}|}{N_{pop}\cdot \max |\mathcal{A}^{(j)}|} \gg 1$ for any $N_{pop}\ge 6$, yet yielding the observed mild, linear increase in runtime. The compression is effective, but because the parallel subspaces still overlap significantly, leading to redundant search that outweighs the benefit of additional subspace search, hence runtime grows slowly rather than shrinking.
    \item Hanoi possesses an inherently sequential structure: the branching factor at each step is small, but the depth is large, making $|\mathcal{A}|$ grow exponentially with horizon. $N_{pop}$ increases, and because $\max |\mathcal{A}^{(j)}|$ shrinks super-linearly (long sequences are split into shorter prefixes), $PSCR(D)$ rises sharply. Consequently, the exponential term $|\mathcal{A}|^{1-\gamma}$ collapses to $\bigl(\frac{|\mathcal{A}|}{N_{pop}}\bigr)^{1-\gamma}$ , producing the 40\% linear speed-up observed from $N_{pop} =1$ to $21$.
\end{itemize}

\paragraph{Subspace size $\ell_{max}$:}
$\ell_{max}$ controls the size of each compressed subspace: a smaller $\ell_{max}$ reduces $\mathcal{A}^{(j)}$ at the cost of approximation fidelity.
\begin{itemize}
    \item In Blocks and Rearrangement, local actions admit compact latent encodings, so $\mathcal{A}^{(j)}$ decreases faster than the compression loss incurred. $PSCR(D)$ improves and runtime drops linearly.
    \item In Hanoi and Grippers, action dependencies in long-horizon require dense latent representations. Shrinking $\ell_{max}$ forces the subspace to contain many distinct action sequences, inflating $\mathcal{A}^{(j)}$ through approximation error (e.g., information loss generated when approximating the original action space with smaller subspaces). Therefore $PSCR(D)$ deteriorates and runtime rises linearly until $\ell_{max}$ falls below the critical threshold ($\approx 11$ for Hanoi), after which the overhead dominates and the exponential bottleneck re-emerges.
\end{itemize}

In summary, the observed runtime of our planner are direct manifestations of $PSCR(D)$. Whenever $PSCR(D)>1$ and grows, the exponential bottleneck is alleviated. Whenever $PSCR(D)\le 1$, the bottleneck reasserts itself, independent of success-rate stability.

\end{document}